\pdfoutput=1
\documentclass[lettersize,journal]{IEEEtran}

\usepackage{graphicx}

\usepackage{amsmath}
\usepackage{array}
\usepackage{booktabs}
\usepackage{multirow}

\usepackage{placeins}
\usepackage{subcaption}

\usepackage{url}
\usepackage{xurl} 
\usepackage{cprotect}
\usepackage[breaklinks=true]{hyperref}
\usepackage{enumitem}

\hypersetup{
    colorlinks=true,
    linkcolor=blue,
    filecolor=magenta,      
    urlcolor=cyan,
}

\newcommand{\eg}{\textit{e.g.},~}
\newcommand{\ie}{\textit{i.e.},~}

\begin{document}

\title{Lowering the Barrier to IREX Participation: Open-Source Algorithms, Toolkit, and Benchmarking for Iris Recognition}

\author{Siamul~Karim~Khan,~\IEEEmembership{Student Member,~IEEE,}
        Patrick~J.~Flynn,~\IEEEmembership{Fellow,~IEEE,}
        and~Adam~Czajka,~\IEEEmembership{Senior~Member,~IEEE}
\thanks{The authors are with the Department of Computer Science and Engineering, University of Notre Dame, Indiana, USA. Corresponding e-mail: aczajka@nd.edu.}%
\thanks{Draft submitted for consideration to {\it IEEE Transactions on Biometrics, Behavior, and Identity Science} (T-BIOM)}%
}

\markboth{Preprint}%
{Khan \MakeLowercase{\textit{et al.}}: IREX-X Open Source Iris Algorithms}

\maketitle

\begin{abstract}
Iris recognition, matured over three decades of development, is now the third principal biometric technique (with face and fingerprint recognition) integrated into critical national security systems. New open-source algorithms are, however, tested exclusively in laboratory settings using datasets with known sample distributions, and usually not sufficiently large to estimate small errors observed for this biometric technique with appropriate statistical guarantees. Large-scale evaluation programs like NIST Iris Exchange (IREX) offer an appealing solution, but they present high barriers to entry because academic algorithms must be written in C++, using a specific API, and adapted to meet strict IREX speed and memory constraints.

The main goal of this paper is to lower these barriers and advance open-source iris recognition large-scale evaluations by offering: (a) two new modern deep learning-based open-source iris matchers (ArcIris and TripletIris), along with their C++ IREX X-compliant implementations, which are the first open-source iris recognition methods included into the IREX X leaderboard (and thus IREX-vetted), as well as new segmentation and iris circular approximation models that can be incorporated into any new iris recognition method, and (b) a performance assessment (according to IREX X testing protocols) of all major and currently available open-source iris recognition solutions.

The paper also provides Python implementations of the new ArcIris and TripletIris methods and discusses the differences one may encounter between C++ and Python implementations of the same conceptually equivalent approaches. Finally, the paper offers open-source, IREX X-compliant C++ implementations of two existing methods: (a) an iris image filtering-based algorithm utilizing human saliency-driven kernels (HDBIF), and (b) a human-interpretable algorithm for detecting and comparing Fuchs' crypts (CRYPTS). In addition to IREX X evaluation results, the paper reports the performance of all methods on major academic benchmarks: Quality-Face/Iris Research Ensemble (Q-FIRE), Warsaw-Biobase Post-Mortem Iris, CASIA-Iris-Thousand-V4, CASIA-Iris-Lamp-V4, IIT Delhi Iris Database, IIITD Contact Lens Iris Database, NDIris3D, and Notre Dame Variable Iris Image Quality Release 2 (VII-Q-R2).
\end{abstract}


\ifCLASSOPTIONpeerreview
\begin{center} \bfseries EDICS Category: 3-BBND \end{center}
\fi

\IEEEpeerreviewmaketitle

\section{Introduction}

Iris recognition technology, known for its accuracy and speed, has become prominent in security applications and has seen an increase in the number of open-source approaches in recent years. However, one still unaddressed concern relates to professional testing of such algorithms, on sequestered and large datasets, guaranteeing fairness and closeness to reality, in which the actual iris samples have never been seen by the developers. Operational aspects such as accuracy, time of execution, and generalization capabilities (to new sensors, subjects, ethnic groups, time between enrollment and verification, etc.) have been evaluated within the Iris Exchange (IREX) program administered by NIST’s Biometrics Research Laboratory \cite{IREX_X}. To date, the many IREX evaluations have been populated largely by commercial (closed-source) algorithm submissions. It is, however, desirable to use the capabilities and high reputation of the IREX program to also incorporate open-source solutions, including those from academic institutions. There are at least three good reasons for stimulating the development of open-source iris recognition methods, and -- if possible -- having them evaluated in programs such as IREX: 

\begin{enumerate}[leftmargin=5.5mm]
    \item[(a)] reproducible, trusted, and professionally-tested algorithms would serve as an important baseline and benchmark for academic efforts to design new iris recognition methods,
    \item[(b)] having an algorithm from academic units submitted to and vetted by IREX X would decrease the reluctance of other academic teams to have their methods evaluated in the IREX program,
    \item[(c)] open-source iris recognition approaches encourage transparency, collaboration, and faster deployments of iris recognition techniques in smaller-scale or pilot implementations before the adoption of professional solutions.
\end{enumerate}

This paper proposes two new open-source, IREX X-vetted iris recognition methods: i) {\it TripletIris}, the ConvNeXt-tiny-based and trained on Triplet loss with Batch-Hard Triplet Mining~\cite{schroff2015facenet}, and ii) {\it ArcIris}, the ResNet100-based and trained with ArcFace loss~\cite{deng2019arcface}. The paper also offers IREX X-compliant open-source versions of two academic iris recognition methods previously developed by the University of Notre Dame: i) an iris image filtering-based algorithm utilizing human saliency-driven kernels (HDBIF \cite{Czajka_WACV_2019b}), and ii) a human-interpretable algorithm for detecting and comparing Fuchs’ crypts (CRYPTS \cite{Chen_TIFS_2016}). The {\bf core novel contributions} of this work are:

\begin{enumerate}[leftmargin=5.5mm]
    \item[(a)] the first open-source iris matchers compliant with the NIST IREX, and already evaluated in the IREX X program for accuracy, search time, template size, and selected demographics effects\footnote{detailed results can be found at \url{https://pages.nist.gov/IREX10/}; {\it TripletIris} = {\ttfamily ndcvrl\_001}, and {\it ArcIris} = {\ttfamily ndcvrl\_002} in the leaderboard},
    \item[(b)] an iris image segmentation approach, incorporating two models: (i) a pixel-wise segmentation model based on a nested U-Net architecture combining residual blocks with shared atrous convolutions, and (ii) a regression model approximating inner and outer iris boundaries directly from the iris image, and
    \item[(c)] two deep neural networks-based iris recognition approaches trained using: (i) triplet loss with batch-hard triplet mining loss, and (ii) ArcFace loss.
\end{enumerate}

The provided C++ codes act as a guide for other researchers developing open-source iris recognition systems to submit their systems to the NIST IREX X evaluation program. Additionally, methodologically equivalent MATLAB and Python codes are provided for those looking for faster adoption of these algorithms in research projects. The repository with all components offered with this paper is available at \url{https://github.com/CVRL/OpenSourceIrisRecognition}. The authors' ambition is to keep this repository well-maintained and expanding to serve as a primary source of various open-source iris recognition tools for the biometrics community.
\section{Existing Open Source Iris Recognition Approaches}

This section attempts to identify all available open-source iris recognition methods, other than those offered in this paper, that were available to the authors at the time of preparing this manuscript. These methods are then evaluated along with the algorithms proposed in this paper in Sec. \ref{sec:performance}.

\subsection{MATLAB Source Code for a Biometric Identification System Based on Iris Patterns}

{\bf Summary:} While no longer serving as the state-of-the-art iris recognition implementation, historically the first open-source algorithm that mimicked Daugman's original approach was developed and released by Masek as his Master's thesis~\cite{Masek_MSc_2003}. The iris recognition process in his implementation involves three standard steps: segmentation (utilizing Hough transforms) to isolate the iris from the eye image and locate eyelids, normalization to account for pupil size variations, and feature encoding to create an iris code (using Log-Gabor filters and phase quantization). Finally, template matching is performed using the fractional Hamming distance metric.

{\bf Source codes:} \url{https://peterkovesi.com/studentprojects/libor/sourcecode.html}

\subsection{Open Source for Iris (OSIRIS)}

{\bf Summary:} 
OSIRIS~\cite{Othman_PRL_2016} is an open-source iris recognition package proposed by the Biosecure Association, with version OSIRISv4.1~\cite{OSIRIS_SDK} serving for a long time as the state-of-the-art open-source iris matcher. A key difference in OSIRISv4.1, compared to a classical Daugman's method, lies in its utilization of the Viterbi algorithm to estimate coarse contours between the pupil, iris, and sclera. This allows for more accurate normalization even with irregular iris and pupil shapes. As in the original Daugman's algorithm, OSIRIS v4.1 utilizes both the real and imaginary components of Gabor wavelets in the feature extraction stage. The fractional Hamming distance is also used here for comparing the iris codes. One technical implementation detail is that OSIRIS is written in C++ and requires the (relatively old) OpenCV 2.4 to be compiled.


{\bf Source codes:}
\url{http://svnext.it-sudparis.eu/svnview2-eph/ref_syst/Iris_Osiris_v4.1} (official). Virtual Machine version of OSIRIS has also been prepared by Clarkson/CiTER (VirtualBox): \url{https://github.com/ClarksonCITeR/osiris_vmbox}

\subsection{University of Salzburg Iris Toolkit (USIT)}

{\bf Summary:} The USIT toolkit~\cite{USIT3} follows a modular approach, allowing for flexibility in combining different algorithms for iris segmentation, encoding, and matching. The combination of USIT tools, applied for evaluations presented in this paper, follows the original authors' recommendation and uses Contrast-Adjusted Hough Transform (CAHT) for iris localization, Quadratic Spline Wavelet (QSW) for feature encoding, and Hamming distance for comparison scoring.

{\bf Source codes:} \url{https://www.wavelab.at/sources/Rathgeb16a/}

\subsection{Triplet Based Iris Recognition without Normalization (ThirdEye)}

{\bf Summary:} The ThirdEye~\cite{Ahmad_BTAS_2019} uses a Convolutional Neural Network (CNN) architecture trained within a triplet-based framework. This approach utilizes segmented and cropped iris images; however, it does not perform an iris normalization step. During training, the model is fed triplets of images: a target image, a positive image belonging to the same class (iris), and a negative image from a different class (iris). The training objective aims to achieve two goals in parallel: minimization of the distance between the network's output vectors for the target and positive samples, and maximization of the distance between the output vectors for the target and negative samples. In the paper, the authors employ a two-stage training process. First, the CNN architecture undergoes pre-training using the standard cross-entropy loss (softmax) where each iris is treated as a separate class. Then, the model is fine-tuned using the triplet-based framework.

{\bf Source codes:} \url{https://github.com/sohaib50k/ThirdEye---Iris-recognition-using-triplets}

\subsection{Dynamic Graph Representation for Partially Occluded Biometrics (DGR)}

{\bf Summary:} CNNs exhibit a significant drop in performance for biometric recognition tasks due to occlusions caused by various factors. The authors propose a novel unified framework, Dynamic Graph Representations (DGR)~\cite{Ren_AAAI_2020}, to address this problem. DGR leverages the strengths of both CNNs and graph models to learn dynamic graph representations specifically for handling occlusions in biometrics. Their framework integrates several recent advancements in computer vision, including attention mechanisms, triplet loss, and graph neural networks, to effectively deal with irregularities due to iris texture occlusions. The authors extend their work to utilize multi-scale features in~\cite{Ren_TPAMI_2023}.

{\bf Source codes:} \url{https://github.com/RenMin1991/Dyamic-Graph-Representation}

\subsection{Human Saliency-Driven Patch-based Matching for Interpretable Post-mortem Iris Recognition (PBM)}

{\bf Summary:} The PBM~\cite{Boyd_WACVW_2023} method was originally devised for forensic applications. Its goal is to assist human examiners by providing interpretable results when matching post-mortem irises. While not originally intended for live irises, there's no inherent limitation to its application in that domain. The PBM method employs a Mask R-CNN model trained to detect iris texture patches corresponding to features annotated by human examiners. As such, these patches don't necessarily correspond to known anatomical landmarks within the iris texture. Following the detection stage, the PBM method utilizes the same human-driven filtering kernels utilized in the HDBIF matcher to encode iris features inside the detected patches. Finally, it calculates similarity scores through hamming distance-based matching of local, geometrically corresponding features, and counting the number of matching feature pairs.

{\bf Source codes:} \url{https://github.com/CVRL/PBM}

\subsection{Worldcoin Iris Recognition Inference System (WCI)}

{\bf Summary:} The WCI~\cite{wldiris} is an open-source iris recognition system developed as part of Worldcoin, a privacy-preserving human identity and financial network, and closely implements the original Daugman's approach \cite{Daugman_TCSVT_2004}. For segmentation, the method utilizes a two-headed autoencoder approach consisting of an encoder that is shared by two decoders: one to annotate pupil, iris, and sclera, and the other to detect non-eye-related elements that can potentially obscure iris texture (\eg eyelashes, hair, etc.)~\cite{lazarski2022two}. Then, it converts the circular iris into a rectangular region using Daugman's rubber sheet model, calculates the iris codes utilizing Gabor wavelets, and considers the phase information when calculating the matching score using fractional Hamming distance. The WCI is the most recent open-source implementation of Daugman's algorithm.

{\bf Source codes:} \url{https://github.com/worldcoin/open-iris}

\section{Datasets Utilized}

\subsection{Segmentation Training Datasets}

%

To train the pixel-wise segmentation model, we utilize a set of iris images with their corresponding ground truth masks, sampled from a large corpus of publicly available datasets. We aimed at achieving a good level of diversity in terms of population, acquisition environments, and potential anomalies, which may be observed in operational scenarios (such as eye conditions or even post-mortem tissue decomposition). This composition was made of samples from the following datasets:

\begin{itemize}[leftmargin=5.5mm]
    \item {\bf BioSec} project baseline corpus \cite{Fierrez_PR_2007} (1200 images), with ground-truth segmentation masks provided by \cite{AlonsoFernandez_IET_2015,Hofbauer_ICPR_2014},
    \item {\bf BATH} \cite{bath_dataset} (148 images), with ground-truth segmentation masks provided by \cite{Trokielewicz_WACV_2020},
    \item {\bf ND-Iris-0405} \cite{Phillips_TPAMI_2010} (1283 images), with ground-truth segmentation masks provided by \cite{Hofbauer_ICPR_2014,Hofbauer_Techreport_2014},
    \item {\bf CASIA-V4-Iris-Interval} \cite{CASIA_Iris_v4_URL} (2639 images), with ground-truth segmentation masks provided by \cite{Hofbauer_ICPR_2014,Hofbauer_Techreport_2014}, 
    \item {\bf UBIRIS v2} \cite{Proenca_TPAMI_2010,UBIRIS_DB_URL} (1923 images; we used only the red channel from these visible-light samples), with ground-truth segmentation masks provided by \cite{Hofbauer_ICPR_2014,Hofbauer_Techreport_2014}, 
    \item {\bf Warsaw-BioBase-Disease-Iris v2.1} \cite{Trokielewicz_IVC_2017}, with ground-truth segmentation masks provided by \cite{Trokielewicz_WACV_2020}, and
    \item {\bf Warsaw-BioBase-Post-Mortem-Iris v3.0} \cite{Trokielewicz_TIFS_2019}, with ground-truth segmentation masks provided by \cite{Trokielewicz_WACV_2020}.
\end{itemize}

To train the model estimating the circular approximations of the iris boundaries, we utilize Open Eye Dataset (OpenEDS)~\cite{garbin2019openeds}. The dataset is collected using a Virtual Reality (VR) head-mounted display. It contains 12,759 images with pixel-level annotations for iris, pupil, and sclera. We filter these images to exclude images where the iris is significantly off-center and utilize these images to train our circle parameter estimation model.

\subsection{Feature Encoding Training Datasets}

To train the proposed feature encoding neural networks for TripletIris and ArcIris, we utilize the bona fide images composed of several publicly available iris datasets released by the University of Notre Dame~\cite{nd_cvrl_datasets} (ND CCL, Photometric Stereo Iris Dataset, ND-Iris-0405 Dataset, ND-GFI, LivDet-Iris 2017, ND Iris Contact Lenses 2010, ND Iris-Template-Aging-2008-2010, ND-TimeLapseIris-2012, ND-CrossSensor-Iris-2013, and ND-Cosmetic-Contact-Lenses-2013). The dataset comprises 286,649 iris images across 2,759 unique identities, where an identity represents a distinct combination of subject and eye (Left or Right).

\subsection{Benchmarks}

To compare different open-source algorithms, we utilize eight datasets, described below. These datasets offer variations across image quality, demographics, and sensors. For convenience, the most important variations are explicitly stressed in a short description of each benchmark below. A copy of the Notre Dame datasets used in this work can be requested at \url{https://cvrl.nd.edu/projects/data/}.

\subsubsection{Quality-Face/Iris Research Ensemble (Q-FIRE)~\cite{johnson2010quality}}
The Q-FIRE dataset investigates the impact of iris image quality on recognition accuracy. This collection includes near-infrared (NIR) videos captured at various distances (5ft, 7ft, 11ft, 15ft, and 25ft), showcasing subjects moving into and out of focus. Additionally, videos were captured of individuals walking through a ``portal'' at 7ft and 15ft to introduce motion blur. These recordings were obtained using a Dalsa 4M30 infrared camera. Baseline iris samples (``MBARK'') were also acquired using an OKI IrisPass EQ5016A sensor. 

\begin{figure}[!htb]
\centering
\includegraphics[width=\linewidth]{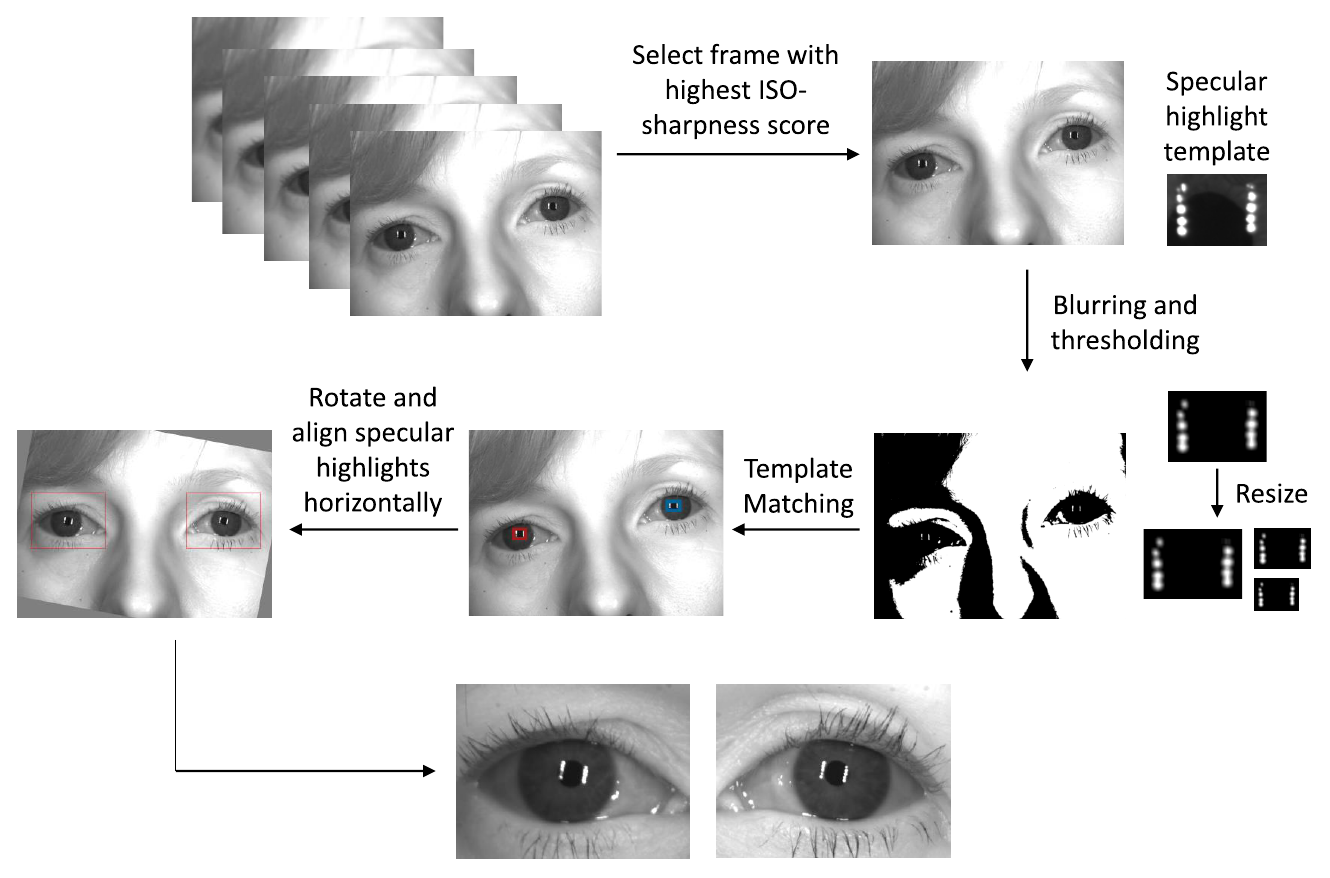}
\caption{Extraction of iris images compliant with ISO/IEC 19794-6 from Q-FIRE videos using a simple template matching approach. The same specular highlight template (shown in the right top corner of this figure) is used to localize the eye region in all images.}
\label{fig:qfire-extraction} 
\end{figure}

{\bf Image variability:} The dataset deliberately incorporates variations in resolution, lighting, focus, gaze/pose angles, motion blur, and occlusions. To ensure that the samples used in our evaluation are ISO-compliant, we implemented a three-step curation process:
\begin{itemize}[leftmargin=5.5mm]
    \item we developed and applied an automated method for extracting ISO-compliant samples from the ``portal'' videos, using deformable template matching (Fig.~\ref{fig:qfire-extraction}),
    \item next, we manually reviewed automatically extracted samples from the step above,
    \item finally, we manually evaluated all MBARK images, retaining only ISO-compliant ones.
\end{itemize}

\subsubsection{Warsaw-Biobase Post-Mortem Iris v3.0 Dataset (WBPMI)~\cite{Trokielewicz_TIFS_2019}}
The WBPMI dataset consists of 1094 near-infrared (NIR) images and 785 visible light images collected from 42 post-mortem subjects and is part of the larger LivDet-Iris 2020 test set~\cite{Das_IJCB_2020, LivDetIris2020}. The NIR images were collected using the commercial iris sensor IriShield M2120U, and color images were collected using a consumer-grade color camera, Olympus TG-3. We utilize only the NIR images for our evaluation purposes. 

{\bf Image variability:} Due to tissue decay, post-mortem irises lose their shape and texture and are harder to segment and recognize, making this dataset a challenging benchmark for testing the robustness of iris recognition against non-ISO-compliant artifacts in images. 

\subsubsection{CASIA-Iris-Thousand V4~\cite{casia_iris_v4}}
The CASIA-Iris-Thousand dataset contains 20,000 iris images from 1,000 subjects, which were collected using an IKEMB-100 dual-eye camera manufactured by IrisKing. During acquisition, the camera’s visual feedback system assisted subjects in adjusting their pose to ensure high-quality capture. As the first publicly available iris dataset containing a full thousand distinct subjects, it serves as an excellent benchmark for studying the uniqueness of iris features and evaluating the scalability of novel iris classification and indexing methods.

{\bf Image variability:} The primary sources of intra-class variations are the presence of eyeglasses and specular reflections. 

\subsubsection{CASIA-Iris-Lamp V4~\cite{casia_iris_v4}}
The CASIA-Iris-Lamp dataset consists of 16,212 iris images from 819 eyes across 411 subjects collected using the handheld OKI IRISPASS-h camera. During acquisition, a proximate lamp was turned on and off to intentionally introduce intra-class variations.

{\bf Image variability:} Due to pupil expansion and contraction under these fluctuating illumination conditions, the iris texture undergoes significant deformations, making the dataset an excellent benchmark for evaluating the impact of iris normalization and the robustness of feature representations.

\subsubsection{IIT Delhi Iris Database (IITD-Iris)~\cite{kumar2010comparison}} The IITD-Iris dataset consists of 2,180 iris images from 218 subjects collected using the JIRIS JPC1000 digital CMOS camera. The images were acquired in an indoor environment at a resolution of $320\times240$ pixels, featuring individuals between the ages of 14 and 55. The dataset was established to evaluate the performance of state-of-the-art recognition systems on a demographic specifically representing Indian users, maintaining all natural variations in image quality as originally acquired. 

{\bf Image variability:} This dataset includes real-world acquisition artifacts without artificial constraints, which makes it an excellent benchmark for evaluating the generalizability of iris segmentation algorithms.

\subsubsection{NDIris3D dataset (ND3D)~\cite{fang2020robust, das2020iris}} The ND3D dataset consists of 6,850 near-infrared (NIR) iris images collected from 176 eyes across 88 subjects. The images were acquired using two distinct commercial sensors: the LG IrisAccess 4000 (LG4000) and the IrisGuard AD 100 (AD100). The collection protocol captured subjects both with and without textured contact lenses from three different manufacturers (Johnson \& Johnson, Ciba Vision, and Bausch \& Lomb). We utilize only the non-PAD (authentic) subset of these images, which provides a noiseless collection of iris images and serves as an excellent benchmark for evaluating the recognition capabilities of our algorithms under ideal acquisition conditions.

{\bf Image variability:} The near-infrared illumination was intentionally varied across multiple angles to enable photometric stereo-based 3D reconstruction of the iris surface. 

\subsubsection{IIITD Contact Lens Iris Database (IIITD-CLI)~\cite{yadav2014unraveling}} The dataset consists of 6,570 iris images from 202 irises (101 subjects), acquired using two distinct devices: the Cogent dual iris sensor (CIS 202) and the VistaFA2E single iris sensor. 

{\bf Image variability:} The collection captures subjects under three conditions: wearing transparent (soft) lenses, wearing textured (color cosmetic) lenses in four different colors, and a baseline without lenses, with products primarily manufactured by CIBA Vision and Bausch \& Lomb. Because our objective is to evaluate the core performance of iris recognition algorithms, we use images with no contact lenses.

\subsubsection{Notre Dame Variable Iris Image Quality Release 2 Non-sequestered Dataset (VII-Q-R2)~\cite{viiq}} The dataset consists of 4,466 near-infrared (NIR) iris images from 203 distinct eyes across 144 subjects and is a specialized partition of the broader VII-Q database prepared to support the National Institute of Standards and Technology (NIST) projects.

{\bf Image variability:} This collection explicitly incorporates a wide spectrum of naturally occurring real-world variations in iris image quality, such as heavily closed eyelids (yet still allowing a human expert to perform iris matching based on a partially visible iris texture), off-axis captured, low exposure or over-exposed samples, low contrast samples, non-ISO-compliant iris placement and size, etc.
\section{Proposed Open Source Methods}

The algorithmic contributions of this work are related to three modules. The first, segmentation module encompasses two models: one estimating the pixel-wise mask, and the second estimating circular approximations of inner and outer iris boundaries. The second and third modules are deep learning-based encoders that return an iris image representation. The subsection below provide details on the proposed methods.

\subsection{Iris Segmentation Module (SEGM)}
\label{sec:segm-module}

\subsubsection{Pixel-Wise Segmentation Model Architecture}
To select the best architecture for our model, we experimented with different modified versions of Nested UNet~\cite{zhou2018unet++}:
\begin{itemize}
\item \textit{NestedResUNet:} Each convolution layer in Nested UNet~\cite{zhou2018unet++} is replaced by a Residual Block~\cite{he2016deep}.
\item \textit{NestedAttentionResUNet:} Each convolution layer in Nested UNet~\cite{zhou2018unet++} is replaced by a Residual Block~\cite{he2016deep} and attention mechanism is introduced into the network~\cite{li2020anu}.
\item \textit{NestedSharedAtrousResUNet:} Each convolution layer in Nested UNet~\cite{zhou2018unet++} is replaced by a Residual Block~\cite{he2016deep} with the convolutions inside the residual block replaced by Kernel-Sharing Atrous Convolutions~\cite{huang2021see}.
\end{itemize}

\begin{figure}[!ht]
\centering
\includegraphics[width=0.95\linewidth]{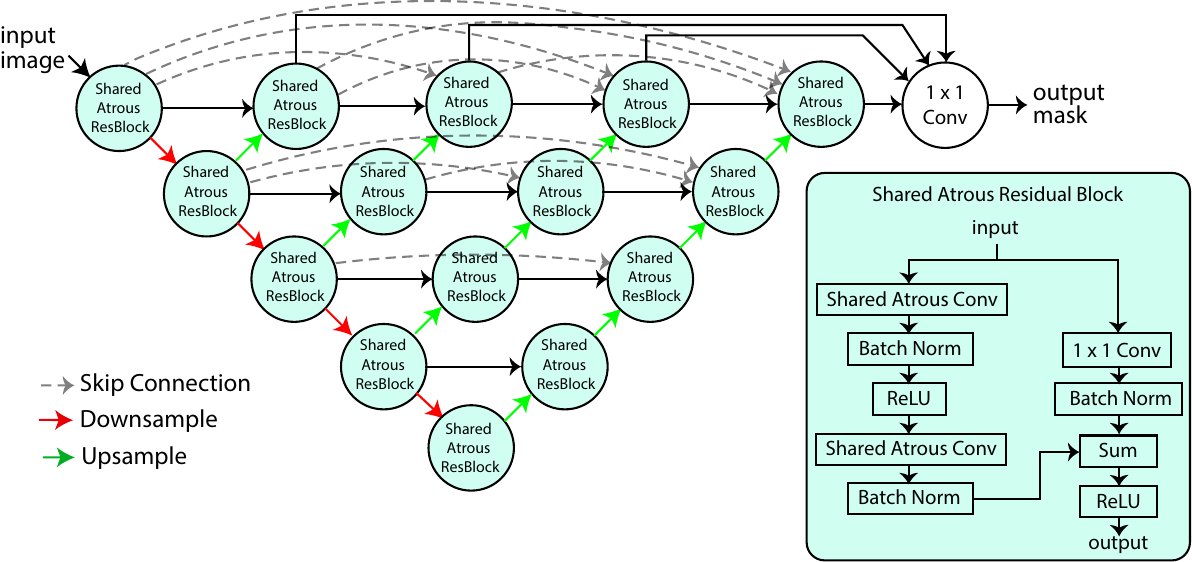}
\caption{Visualization of the {\it NestedSharedAtrousResUNet} architecture used in the segmentation model.}
\label{fig:nsarunet}
\end{figure}

We provide an Intersection over Union (IoU) and runtime comparison for the different architectures in Table~\ref{tab:ioucompseg}, according to which the {\it NestedSharedAtrousResUNet} architecture, presented in Figure~\ref{fig:nsarunet}, performs best. However, due to time constraints enforced by the IREX X program, we utilize the {\it NestedSharedAtrousResUNet} with a starting width of 16 channels in the IREX X submissions. 

\begin{table}[!ht]
\centering
\caption{IoU and runtime for segmentation models. The runtime corresponds to the CPU (AMD Ryzen 7 5800H) inference by the C++ implementation, averaged over 100 runs.}
\label{tab:ioucompseg}
\begin{tabular}{l|c|c}
\toprule
\textbf{Network}      & \textbf{Validation} & \textbf{Runtime} \\ 
\textbf{Architecture}      & \textbf{IoU} & \textbf{(ms)} \\ \midrule 
NestedResUNet (w = 16)             & 0.938251  &  130.50\\ \hline
NestedResUNet (w = 32)             & 0.938770  &  390.08\\ \hline
NestedAttentionResUNet (w = 16)    & 0.904661  &  184.62\\ \hline
NestedAttentionResUNet (w = 32)    & 0.938646  &  493.18\\ \hline
NestedSharedAtrousResUNet (w = 16) & 0.938446 & 153.75\\ \hline
NestedSharedAtrousResUNet (w = 32) & 0.939373  &   384.11\\ 
\bottomrule
\end{tabular}
\end{table}

\subsubsection{Circular Approximation Model Architecture}

The second segmentation model estimates the inner and outer iris boundaries using circular approximations, as in the original Daugman's approach~\cite{Daugman_TCSVT_2004}. We tested three different backbones, two versions of ResNet~\cite{he2016deep}: ResNet18 and ResNet34, and the smallest version of ConvNeXt~\cite{liu2022convnet}: ConvNeXt-tiny. Table~\ref{tab:ioucompcirc} shows the IoU and runtime obtained for these models. Again, due to the processing time constraints enforced by the IREX X program, we utilize the model with the ResNet18 backbone in the IREX X submissions. 

\begin{table}[!ht]
\centering
\caption{IoU and runtime obtained for different backbones used in the circular approximation model. The runtime corresponds to the CPU (AMD Ryzen 7 5800H) inference by the C++ implementation, averaged over 100 runs.}
\label{tab:ioucompcirc}
\begin{tabular}{l|c|c}
\toprule
\textbf{Network}      & \textbf{Validation} & \textbf{Runtime} \\ 
\textbf{Architecture}      & \textbf{IoU} & \textbf{(ms)} \\ \midrule 
ResNet18             & 0.936074  &   30.39     \\ \hline
ResNet34             & 0.936114  &   44.83     \\ \hline
ConvNeXt-tiny        & 0.938020  &   77.34     \\
\bottomrule
\end{tabular}
\end{table}

To estimate the circular parameters from the output of the backbone, we perform a $1\times1$ convolution to reduce the output to 6 channels. Then, we have a separate fully-connected layer per channel, \ie we flatten each channel and pass it through a fully-connected layer to get 6 output nodes for a total of 36 output nodes. After this, we have a linear layer to get 6 output nodes representing the 6 parameters: pupil circle center ($px, py$), pupil circle radius ($pr$), iris circle center ($ix, iy$), and iris circle radius ($ir$).

Circular iris boundaries were estimated from the segmentation masks using the Hough transform. These estimates then served as pseudo-ground truth targets to train the circular approximation regression model.

\subsubsection{Losses} We utilize a combination of two different losses to train the pixel-wise segmentation model: Cross-Entropy Loss and generalized DICE Loss~\cite{sudre2017generalised}. The Cross-Entropy Loss is defined as:
\begin{equation}
\mathcal{L}_\text{CE} = y log (\hat{y}) + (1 - y) log (1 - \hat{y})
\end{equation}
while the generalized DICE loss is defined as:
\begin{equation}
\mathcal{L}_\text{DICE} = 1 - \frac{2y\hat{y} + 1}{y + \hat{y} + 1}
\end{equation}
where $y$ is the ground truth mask and $\hat{y}$ is the predicted confidence map. Our overall loss is the summation of these two losses:
\begin{equation}
\mathcal{L} = \mathcal{L}_\text{CE} + \mathcal{L}_\text{DICE}
\end{equation}

To train the circular approximation model, we utilize the smooth L1 loss~\cite{girshick2015fast}, defined as:
\begin{equation}
\mathcal{L}_\text{param} = \sum_{v \in \{px, py, pr, ix, iy, ir\}}\text{Smooth}_{L1}(v - \hat{v})
\end{equation}
where $\hat{v}$ represents the value predicted by the model for $v$ and $\text{Smooth}_{L1}(x)$ is defined as:
\begin{equation}
\text{Smooth}_{L1}(x) = 
\begin{cases}
      0.5 x^2 & \text{if $|x| < 1$}\\
      |x| - 0.5 & \text{otherwise}\\
\end{cases} 
\end{equation}

\begin{figure*}[!ht]
\centering
\includegraphics[width=\linewidth]{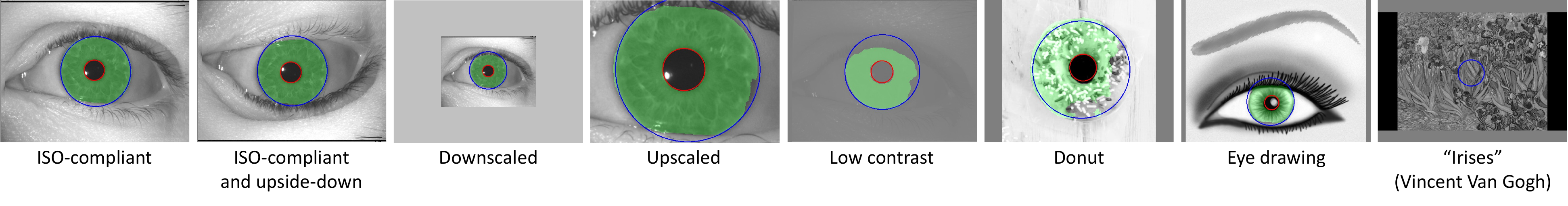}
\caption{Sample images from the NIST IREX X validation dataset, shown with overlaid segmentation masks and iris circular boundaries estimated by the trained models.}
\label{fig:segm_nist}
\end{figure*}

Figure~\ref{fig:segm_nist} visualizes pixel-wise segmentations and estimation of iris boundaries on sample images from the NIST IREX validation dataset, used to verify recognition methods before generating the submission package for formal evaluation.

\subsection{\texorpdfstring{TripletIris: ConvNeXt trained with Batch-Hard Triplet Loss}{TripletIris: ConvNeXt trained with Batch-Hard Triplet Loss}}

The iris image encoding model employed in this configuration is ConvNeXt-tiny \cite{liu2022convnet}, the most compact variant of the ConvNeXt family, trained on polar-normalized iris images using a batch-hard triplet loss. This specific model size was selected to satisfy the inference timing constraints enforced by the IREX X program.

Triplet loss~\cite{schroff2015facenet} optimizes the embedding space by enforcing a structural constraint: the distance between an anchor image and a negative sample (an image from a different identity) must exceed the distance between the anchor and a positive sample (an image of the same identity) by a predefined margin. This mechanism effectively clusters intra-class data points together while pushing inter-class data points apart.

To ensure that our model learns from the triplets effectively, we utilize batch-hard triplet mining. This strategy targets only the most difficult cases within a mini-batch, \ie the furthest positive sample and the closest negative sample for a given anchor. Focusing on these boundary cases forces the model to learn more robust and highly discriminative feature representations at a much lower cost than focusing on all possible triplets.

\subsection{\texorpdfstring{ArcIris: ResNet100 trained with ArcFace Loss}{ArcIris: ResNet100 trained with ArcFace Loss}}

The encoding model employed in this configuration is a 100-layer Residual Network (ResNet100)~\cite{he2016deep} trained on polar-normalized iris images using Additive Angular Margin Loss (ArcFace)~\cite{deng2019arcface}. We switch from ConvNeXt-tiny to ResNet100 because we find that ResNet100 offers a considerable speedup during inference without any drop in recognition performance compared to the ConvNeXt model.

ArcFace optimizes embeddings in the angular domain. It normalizes both the feature vectors and the weights of the final fully connected layer, projecting the representations onto a unit hypersphere. During training, a fixed additive angular margin is applied directly to the angle between the embedded feature vector and its corresponding target class weight center. This additive margin strictly penalizes the classification loss, forcing the model to compress intra-class variance while simultaneously expanding inter-class separation.

The primary drawback of Triplet Loss (applied in TripletIris matcher) is its dependency on sampling strategies. Because there is a combinatorial explosion of $O(N^3)$ possible triplets in a dataset, the model cannot efficiently learn from all of them. Thus, it requires complex mining techniques (like the batch-hard strategy) to find discriminative samples. If the mined triplets are too easy, the model stops learning; if they are too hard, the gradients can become noisy and destabilize training. However, ArcFace bypasses this issue by computing the loss against fixed class weight centers rather than relative sample pairs. Thus, ArcFace requires no sample mining, leading to significantly more stable and efficient training that usually leads to better biometric recognition performance.

\section{IREX-X-Compliant Implementation Details}
This section outlines the critical design choices required to port the proposed methods to the C++ NIST IREX X API and highlights the programmatic differences between implementations. In addition to two new proposed open-source IREX-X-vetted iris recognition methods, we also release IREX-X-compliant C++ adaptations of HDBIF~\cite{Czajka_WACV_2019b} and CRYPTS~\cite{Chen_TIFS_2016} iris matchers.

\subsection{Adapting Methods for NIST IREX X}

Integrating an algorithm into the NIST IREX X API demands wrapping algorithmic logic inside a strict C++ interface. The API dictates rigorous data structures for input images and output templates, precise memory management, and strict threading controls. This section outlines the critical steps for a successful translation.

\subsubsection{Threading Constraints and Inference Optimization} The NIST validation routine achieves concurrency by launching multiple independent processes via \texttt{fork()} rather than intra-process multi-threading. To prevent severe CPU thread contention and evaluation timeouts, implementations must explicitly disable multi-threading across all underlying libraries. We enforce this by invoking \texttt{at::set\_num\_threads(1)} and \texttt{at::set\_num\_interop\_threads(1)} for LibTorch \cite{LibTorch}, and \texttt{cv::setNumThreads(0)} for OpenCV \cite{OpenCV}. 

When deploying PyTorch models via the LibTorch C++ API, disabling gradient tracking is mandatory. Inference calls must be wrapped within \texttt{torch::AutoGradMode enable\_grad(false)} and \texttt{c10::InferenceMode guard(true)}. Neglecting this allows computational graphs to silently accumulate gradients, causing out-of-memory (OOM) failures when matching millions of templates. 

\subsubsection{Porting PyTorch Models to LibTorch for C++} 

At the time of developing our C++ IREX-X-compliant methods, Pytorch provided serialization support via TorchScript to convert Python-native models into compiled \texttt{.pt} models to run natively on the C++ backend. This was achieved through either \textit{Tracing} or \textit{Scripting}. 

\begin{itemize}[leftmargin=5.5mm]
    \item \textit{Tracing:} Passes dummy inputs through the model, freezing the executed operations into a static graph. However, tracing ignores data-dependent dynamic control flow; unexecuted branches are lost, risking catastrophic evaluation errors if those paths are later required. 
    \item \textit{Scripting:} Parses the Python abstract syntax tree (AST) to preserve dynamic logic. This imposes strict typing constraints, requiring developers to add type hints (\eg \texttt{@torch.jit.script}), strip Pythonic idioms, and register custom operators. 
\end{itemize}

Because our neural networks lack data-dependent branches, we exclusively utilize {\it Tracing} for our implementations. Finally, to accelerate the extraction and identification phases, we load all static configurations (YAML/JSON) and network weights into memory in the initialization function.

\subsubsection{Image Preprocessing} 

The \texttt{createIrisTemplate} method in the IREX X API is provided with raw image buffers. Because the sequestered IREX X data lacks strict resolution and color depth constraints, preprocessing is important. We proactively verify color depth (24-bit RGB vs.\ 8-bit grayscale), extract the luminance channel via \texttt{cv::split}, enforce a standard aspect ratio ($4:3$) via padding, and scale to the final input resolution ($320 \times 240$ for the segmentation and circular approximation models, and $640 \times 480$ for the iris image normalization function). 

\subsubsection{Iris Image Quality Checks}
\label{sec:ImageQualityChecks}

We recommend filtering out low-quality iris images. Instead of more restrictive ISO/IEC 29794-6 quality metrics, in the implementation offered with this paper, we only reject images showing obvious biological deviations from a normal iris, based on three parameters: iris radius $r_i$, pupil radius $r_p$, and the number of iris pixels $A_{iris}$ given by the segmentation mask, namely:
\begin{itemize}[leftmargin=5.5mm]
\item \textbf{Abnormal radii:} reject if $r_i \le r_p$;
\item \textbf{Insufficient radii:} reject if $r_p \le 12$ px or $r_i \le 16$ px;
\item \textbf{Abnormal pupil-to-iris ratio ($\alpha$):} reject if 

$$\alpha = \frac{r_p}{r_i} \notin [0.1, 0.8];$$

\item \textbf{Insufficient iris visible:} reject if the amount of iris visible is too low, that is:
$$\frac{A_{iris}}{\pi {(r_i + r_p)(r_i - r_p)}} < 0.1;$$

\item \textbf{Excessive concentric deviation:} reject if the Euclidean distance between the pupil center $(x_p, y_p)$ and iris center $(x_i, y_i)$ exceeds half of the iris radius, that is: 
$$ \frac{\sqrt{(x_p - x_i)^2 + (y_p - y_i)^2}}{r_i} > 0.5.$$
\end{itemize} 

\subsubsection{Template Serialization} 
IREX X API treats biometric templates as contiguous byte arrays (\texttt{std::vector<uint8\allowbreak\_\allowbreak t>}). The \texttt{FRVT::Image::IrisLR} eye-type enumeration must be cast to a \texttt{uint8\allowbreak\_\allowbreak t} and prepended as the first byte, enabling the matcher to filter out cross-eye comparisons (left vs.\ right) during 1:N searches. Multi-dimensional templates must be serialized flat to guarantee correct memory access during matching. We handle three distinct serialization scenarios:
\begin{itemize}[leftmargin=5.5mm]
\item \textbf{Floating-Point Vectors (ArcIris):} Deep learning embeddings (\texttt{std::vector<double>}) are cast via \texttt{reinterpret\allowbreak\_\allowbreak cast<uint8\allowbreak\_\allowbreak t*>}, scaling the size by \texttt{sizeof(double)}.
\item \textbf{Binary Tensors (HDBIF):} 3D binary codes and 2D occlusion masks generated by LibTorch are extracted via \texttt{tensor.data\allowbreak\_\allowbreak ptr()} and copied sequentially.
\item \textbf{Morphological Matrices (CRYPTS):} Spatial features stored in 2D \texttt{cv::Mat} objects are pushed byte-by-byte into the vector.
\end{itemize}

\subsubsection{Identification and Search Logic} 
In operational databases, eye types are frequently labeled as {\it Unspecified}. If a probe is {\it Unspecified}, it must be evaluated against {\it Left}, {\it Right}, and {\it Unspecified} gallery templates, dynamically returning the minimum dissimilarity score. 

Scores are then encapsulated into \texttt{FRVT\_1N::Candidate} structures. Candidates must be ranked by similarity or dissimilarity, sorted, and truncated to the required \texttt{candidateListLength}. The underlying logic must also reliably aggregate comparisons when the IREX validation routine provides multiple iris images ({\it Left} eye, {\it Right} eye, or {\it Unspecified}) for the same identity. In our implementation, {\it Unspecified} iris images from one identity are matched against both the {\it Left} and {\it Right} templates of the paired identity. Additionally, we perform strictly homologous comparisons (\ie {\it Left} vs. {\it Left}, and {\it Right} vs. {\it Right}). The final aggregated comparison score for the comparison pair is computed as the median of all individual comparison scores.

\subsubsection{Time Constraints} 
IREX X strictly caps template creation at 1.5 seconds for $640 \times 480$ iris images. Additionally, a 1:N search against 500,000 templates (returning 50 candidates from both eyes) must conclude within 25 seconds.

To reserve computational headroom for the other algorithms, we deploy lightweight architectures for segmentation and circle parameter estimation (as described in Sec. \ref{sec:segm-module}).

ArcIris and TripletIris use compact networks (ResNet100 and ConvNeXt-tiny). Their matching phases use fast Euclidean or Cosine distance vector comparisons, comfortably clearing search time limits.

HDBIF extraction relies on convolutions, easily meeting the 1.5-second template creation time constraint. However, its search phase calculates fractional Hamming distance via bitwise XOR operations across multiple spatial shifts. To meet the 25-second limit, we optimize the search by initially evaluating only even shifts, and the $\pm 1$ shifts are subsequently evaluated only adjacent to the optimal even shift.

Finally, the CRYPTS method implementation fails the time constraints, especially for matching, as it requires solving a computationally expensive linear programming problem for the two-dimensional Earth Mover's Distance. It is offered as an IREX-X-compliant implementation due to its potential value as the only known to us human-interpretable open source iris matcher, for which relaxing the time constraints or limiting the image gallery size may be applied in the future.

\subsection{Programming Language-Sourced Discrepancies}

For {\it ArcIris}, {\it TripletIris}, and {\it HDBIF}, we observed minor variations between C++ and Python implementations. They are statistically insignificant, but we found it potentially useful to discuss them in this paper, since such mathematical artifacts caused by shifting across libraries, memory architectures, and compilers are rarely discussed in research papers.

\subsubsection{Preprocessing Variations} The Python implementations use Python Imaging Library (PIL) \texttt{Image.resize} function to scale and pad images, whereas the C++ adaptations utilize OpenCV routines (\texttt{INTER\_LINEAR}). Because these libraries implement bilinear interpolation with subtle differences in coordinate mapping and sub-pixel rounding, minor pixel-level discrepancies emerge in the Cartesian and subsequent polar ``rubbersheet'' images before feature extraction.

\subsubsection{Feature Extraction Variations} Transitioning from Python's PyTorch to C++'s LibTorch introduces minor floating-point accumulation differences during matrix operations, despite identical network weights in models like ResNet18 and ConvNeXt.

\subsubsection{Matching Variations} {\it ArcIris} and {\it TripletIris} compute distances natively in C++ using \texttt{double} precision accumulators, replacing Python's \texttt{NumPy} and \texttt{math} libraries. For {\it HDBIF}, \texttt{NumPy}'s \texttt{np.roll} is replaced by LibTorch's \texttt{at::roll} to shift boolean arrays for fractional Hamming distance calculations.

{\it HDBIF} exhibits slightly higher variation than {\it ArcIris} and {\it TripletIris}. The continuous floating-point embeddings of {\it ArcIris}/{\it TripletIris} render fractional changes negligible. In contrast, {\it HDBIF} hard-thresholds convolution responses at zero to generate binary code, and a slight convolution difference for results close to zero can trigger bit flips (these bits are sometimes referred to as \textit{fragile bits} \cite{Hollingsworth_TPAMI_2011}). Furthermore, {\it ArcIris} and {\it TripletIris} models are translation-invariant, whereas {\it HDBIF}'s raw texture filters are highly sensitive to the sub-pixel shifts caused by the PIL vs.\ OpenCV interpolation differences.

\begin{figure*}[!htbp]
    \vskip1mm
    \centering
    \begin{subfigure}[b]{0.24\textwidth}
        \centering
        \includegraphics[width=\textwidth]{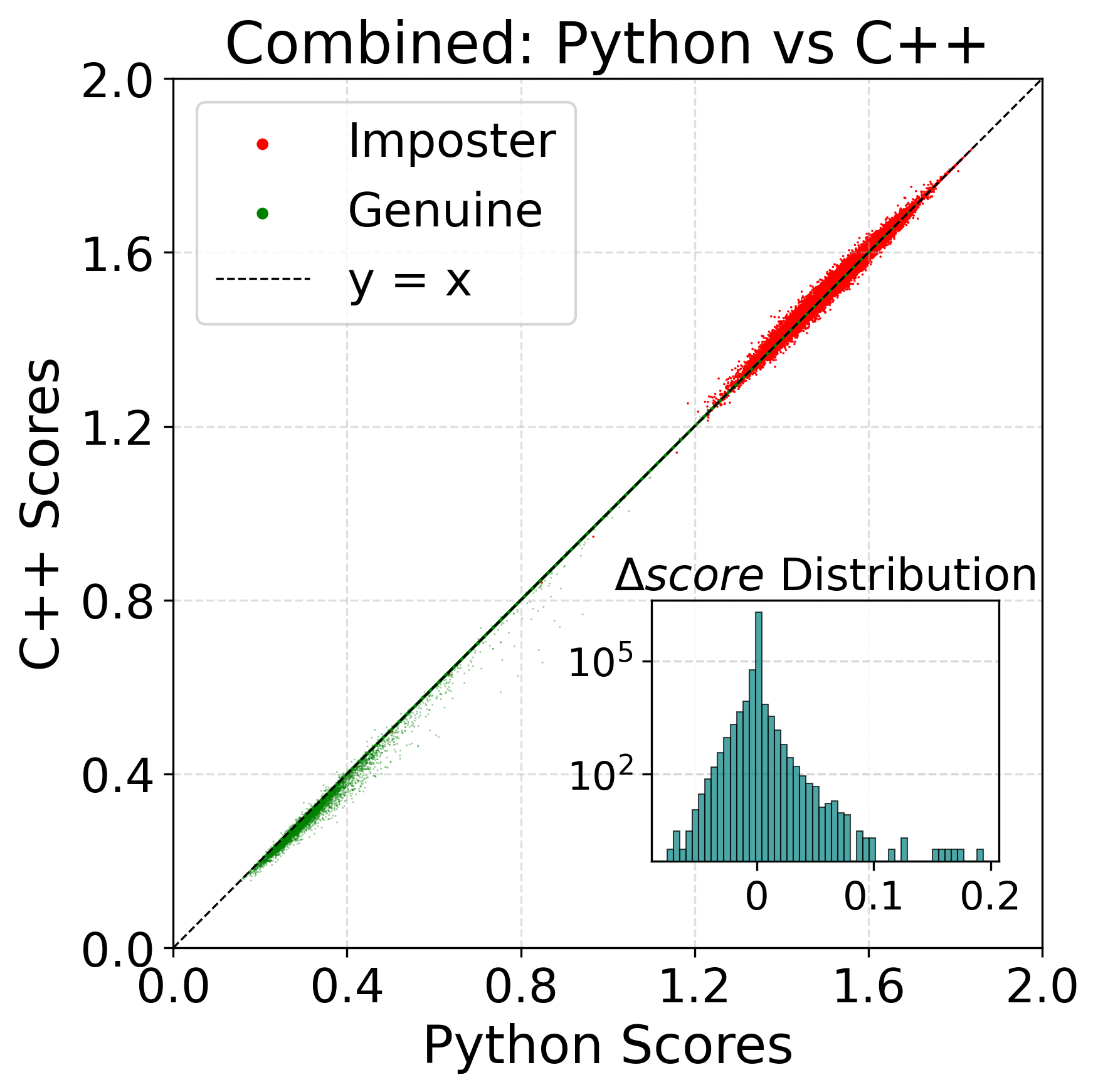}
        \caption{ArcIris}
        \label{fig:arciris_scatter}
    \end{subfigure}
    \hfill 
    \begin{subfigure}[b]{0.235\textwidth}
        \centering
        \includegraphics[width=\textwidth]{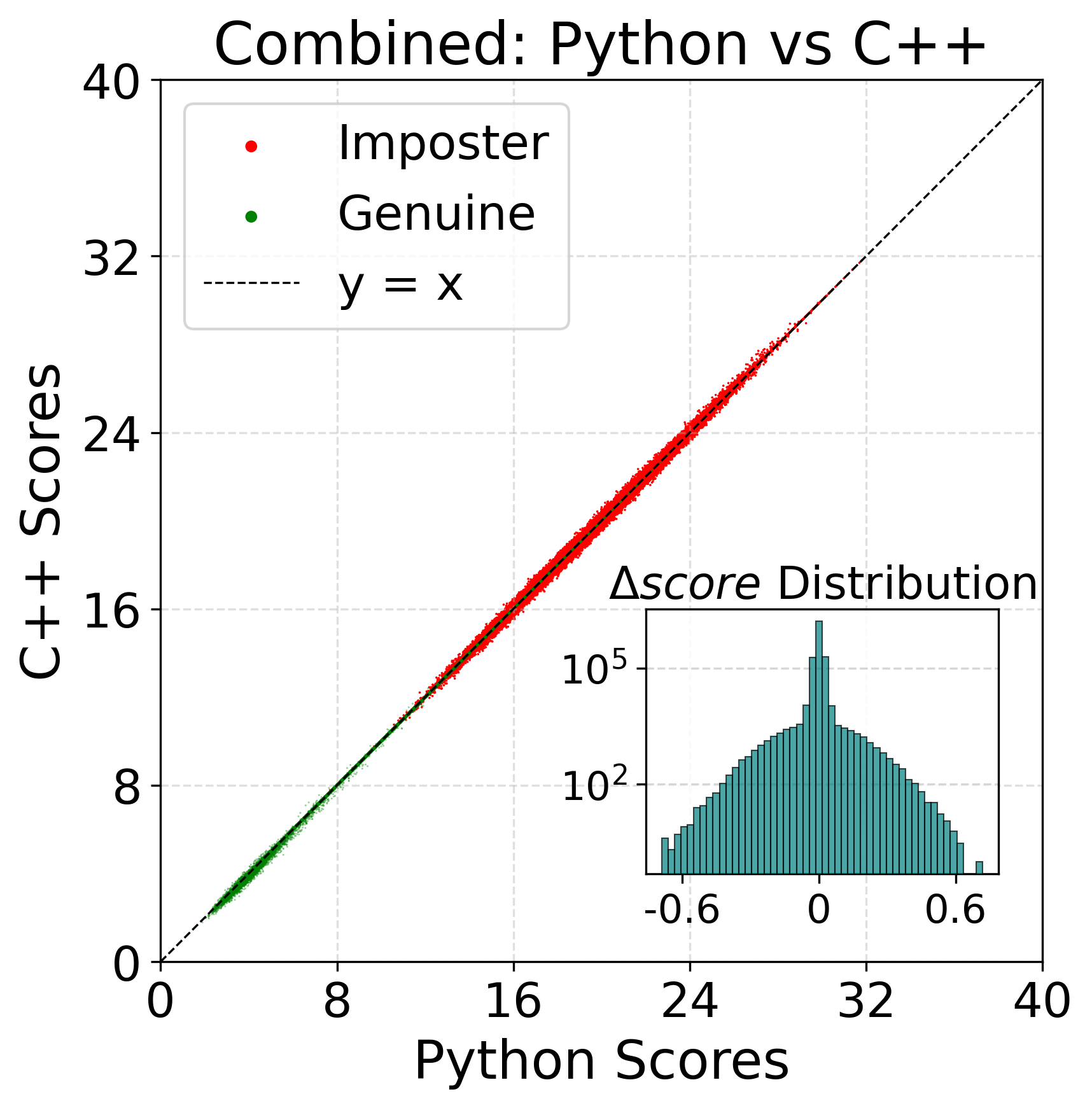}
        \caption{TripletIris}
        \label{fig:tripletiris_scatter}
    \end{subfigure}
    \hfill
    \begin{subfigure}[b]{0.249\textwidth}
        \centering
        \includegraphics[width=\textwidth]{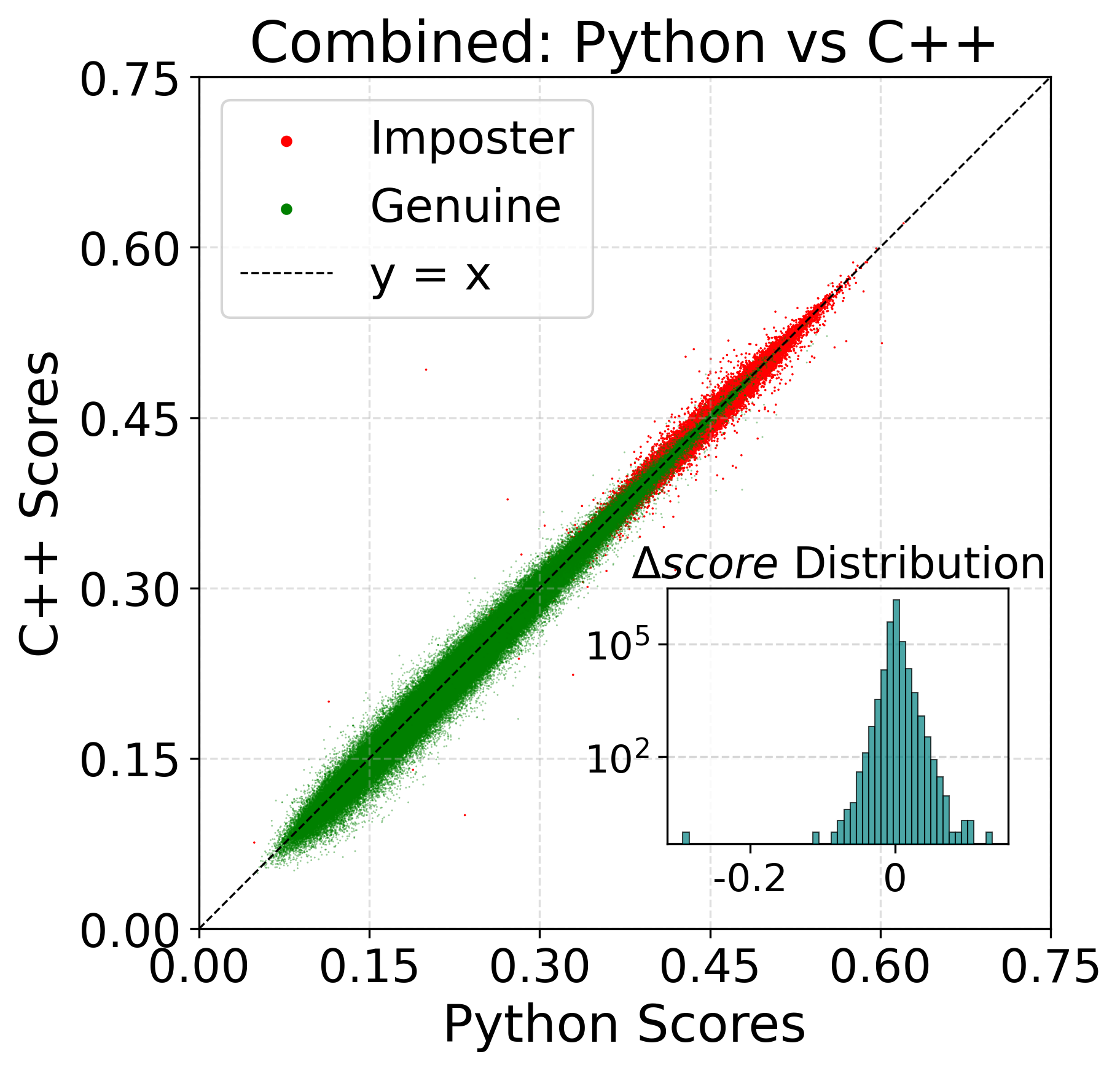}
        \caption{HDBIF}
        \label{fig:hdbif_scatter}
    \end{subfigure}
    \hfill
    \begin{subfigure}[b]{0.249\textwidth}
        \centering
        \includegraphics[width=\textwidth]{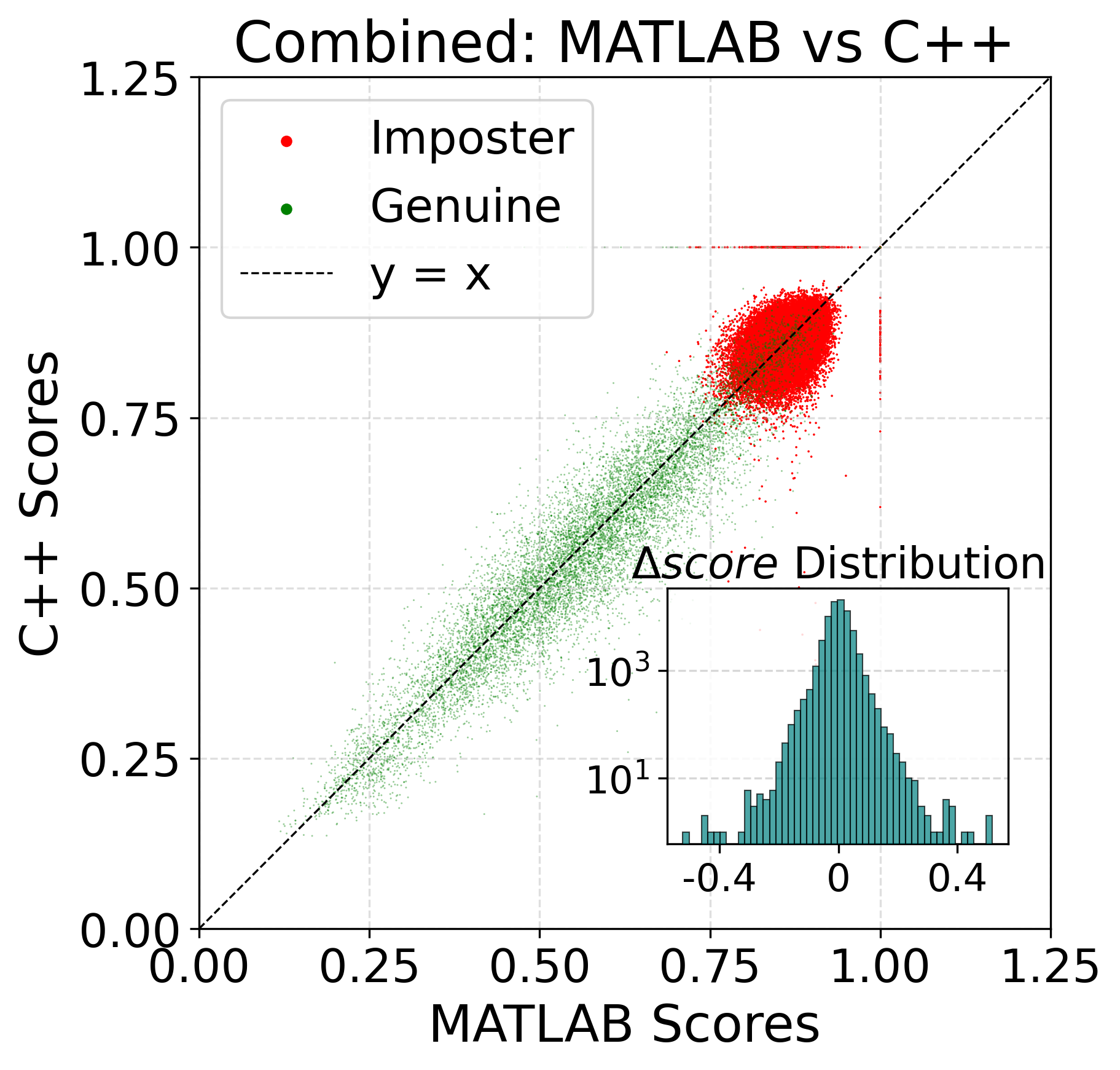}
        \caption{CRYPTS}
        \label{fig:crypts_scatter}
    \end{subfigure}
    
    \caption{Scatter plots juxtaposing comparison scores generated by the IREX X C++ implementation (x-axis) and the Python/MATLAB implementation (y-axis), across all of the evaluation datasets. The tight clustering along the diagonal, as well as a strong peak of differences ($\Delta$) between comparison scores at $0$, indicates high consistency between the implementations.}
    \label{fig:main_figure}
\end{figure*}

The parity columns in Tables~\ref{tab:hdbif_parity}, \ref{tab:arciris_parity}, and \ref{tab:tripletiris_parity} summarize the differences (via Mean Absolute Difference, MAD), which are low, and the agreement ($R^2$), which is high ($R^2 \equiv 1.0$) of the comparison scores between C++ and Python implementations for {\it HDBIF}, {\it ArcIris} and {\it TripletIris}, respectively. Additionally, $\Delta$ plots in Figures~\ref{fig:arciris_scatter}, \ref{fig:tripletiris_scatter}, and \ref{fig:hdbif_scatter} demonstrate tight clustering along the diagonal and a strong $\Delta$ score peak at zero. Thus, our cross-implementation validation suggests that the C++ adaptations faithfully preserve the biometric performance of their open-source Python counterparts.

The discrepancies for the {\it CRYPTS} method, which was ported from Matlab to IREX X C++, are larger. The original implementation relies on Mathworks proprietary functions (\texttt{bwareopen}, \texttt{bwconncomp}, \texttt{imreconstruct}, \texttt{imfill}, and \texttt{linprog}). We replicated these functionalities in C++ primarily using OpenCV functions. Because the underlying algorithms differ, substantial divergence exists between the codebases (see Table~\ref{tab:crypts_parity}). For instance, \texttt{bwareopen} and \texttt{bwconncomp} were mapped to OpenCV's \texttt{connectedComponentsWithStats} utilizing the Spaghetti Algorithm~\cite{bolelli2019spaghetti}, while \texttt{imreconstruct} uses Vincent's algorithm~\cite{vincent1993morphological}. 

Most notably, the C++ linear programming solver HiGHS~\cite{huangfu2018parallelizing} yields negative $R^2$ values for impostor scores. As seen in Figure~\ref{fig:crypts_scatter}, the 2D Earth's Mover Distance (EMD) used in {\it CRYPTS} matching defaults to a 1.0 score under two failure conditions: either it fails an initial pre-check (occurring when the compared input images differ excessively in size or lack sufficient overlap), or the linear programming setup required to solve the EMD problem fails. Because MATLAB and C++ trigger these failures at different instances, distinct $x = 1.0$ and $y = 1.0$ lines appear in the scatter plot, especially for the impostor scores. Ultimately, however, the C++ implementation achieves similar biometric performance to the MATLAB original implementation.

To manage the substantial computational overhead of {\it CRYPTS}, we conducted our evaluations of this method using data sourced from a representative subset of 10\% identities from each dataset.

\begin{table*}[t]
\centering
\caption{{\it HDBIF} Cross-Implementation Parity and Biometric Performance (Python vs. IREX X C++)}
\label{tab:hdbif_parity}
\resizebox{\textwidth}{!}{
\begin{tabular}{lcccccccccccc}
\toprule
\multirow{2}{*}{\textbf{Dataset}} & \multicolumn{2}{c}{\textbf{EER (\%)}} & \multicolumn{2}{c}{\textbf{FNMR (\%) @ 0.1\% FMR}} & \multicolumn{2}{c}{\textbf{FNMR (\%) @ 0.01\% FMR}} & \multicolumn{3}{c}{\textbf{Genuine Scores Parity}} & \multicolumn{3}{c}{\textbf{Impostor Scores Parity}} \\
\cmidrule(lr){2-3} \cmidrule(lr){4-5} \cmidrule(lr){6-7} \cmidrule(lr){8-10} \cmidrule(lr){11-13}
 & \textbf{Python} & \textbf{C++} & \textbf{Python} & \textbf{C++} & \textbf{Python} & \textbf{C++} & \textbf{MAD} & \textbf{Max $\Delta$} & \textbf{$R^2$} & \textbf{MAD} & \textbf{Max $\Delta$} & \textbf{$R^2$} \\
\midrule
CASIA-Iris-Lamp & 1.5164 & 1.4313 & 3.3004 & 3.1151 & 6.5047 & 6.3199 & 0.0072 & 0.0672 & 0.9723 & 0.0022 & 0.0653 & 0.9888 \\
CASIA-Iris-Thousand & 3.3814 & 3.2566 & 11.2047 & 10.7664 & 24.3348 & 24.4664 & 0.0072 & 0.0963 & 0.9782 & 0.0022 & 0.2923 & 0.9891 \\
IITD-Iris & 0.1423 & 0.1360 & 0.1665 & 0.1665 & 2.2898 & 2.6644 & 0.0132 & 0.0731 & 0.8041 & 0.0020 & 0.0273 & 0.9822 \\
ND3D & 0.3991 & 0.3744 & 0.8518 & 0.7593 & 1.9278 & 1.9042 & 0.0059 & 0.0451 & 0.9821 & 0.0016 & 0.0150 & 0.9922 \\
WBPMI & 6.6983 & 6.5822 & 17.9420 & 17.7483 & 29.3681 & 28.2268 & 0.0038 & 0.0432 & 0.9953 & 0.0017 & 0.0347 & 0.9923 \\
Q-FIRE & 1.1865 & 1.1629 & 2.2339 & 2.1932 & 3.6352 & 3.5415 & 0.0044 & 0.0448 & 0.9877 & 0.0017 & 0.0519 & 0.9935 \\
\midrule
\textbf{Combined} & 4.0334 & 3.9619 & 10.4528 & 10.2624 & 18.7377 & 18.4642 & 0.0059 & 0.0963 & 0.9867 & 0.0021 & 0.2923 & 0.9915 \\
\bottomrule
\end{tabular}
}
\end{table*}

\begin{table*}[t]
\centering
\caption{{\it ArcIris} Cross-Implementation Parity and Biometric Performance (Python vs. IREX X C++)}
\label{tab:arciris_parity}
\resizebox{\textwidth}{!}{
\begin{tabular}{lcccccccccccc}
\toprule
\multirow{2}{*}{\textbf{Dataset}} & \multicolumn{2}{c}{\textbf{EER (\%)}} & \multicolumn{2}{c}{\textbf{FNMR (\%) @ 0.1\% FMR}} & \multicolumn{2}{c}{\textbf{FNMR (\%) @ 0.01\% FMR}} & \multicolumn{3}{c}{\textbf{Genuine Scores Parity}} & \multicolumn{3}{c}{\textbf{Impostor Scores Parity}} \\
\cmidrule(lr){2-3} \cmidrule(lr){4-5} \cmidrule(lr){6-7} \cmidrule(lr){8-10} \cmidrule(lr){11-13}
 & \textbf{Python} & \textbf{C++} & \textbf{Python} & \textbf{C++} & \textbf{Python} & \textbf{C++} & \textbf{MAD} & \textbf{Max $\Delta$} & \textbf{$R^2$} & \textbf{MAD} & \textbf{Max $\Delta$} & \textbf{$R^2$} \\
\midrule
CASIA-Iris-Lamp & 0.7018 & 0.7024 & 1.1863 & 1.1903 & 1.8367 & 1.8406 & 0.0004 & 0.0032 & 1.0000 & 0.0003 & 0.0024 & 1.0000 \\
CASIA-Iris-Thousand & 1.3202 & 1.3213 & 3.3890 & 3.3990 & 5.7950 & 5.8285 & 0.0005 & 0.0039 & 1.0000 & 0.0004 & 0.0039 & 0.9999 \\
IITD-Iris & 0.0510 & 0.0425 & 0.0833 & 0.0833 & 1.2081 & 0.9581 & 0.0144 & 0.1935 & 0.9721 & 0.0078 & 0.0766 & 0.9810 \\
ND3D & 0.1713 & 0.1713 & 0.2203 & 0.2203 & 0.3856 & 0.3856 & 0.0001 & 0.0005 & 1.0000 & 0.0000 & 0.0004 & 1.0000 \\
WBPMI & 8.8369 & 8.8416 & 25.5732 & 25.5599 & 36.9348 & 36.9048 & 0.0003 & 0.0027 & 1.0000 & 0.0002 & 0.0024 & 1.0000 \\
Q-FIRE & 0.6856 & 0.6828 & 0.9315 & 0.9283 & 1.2763 & 1.2763 & 0.0004 & 0.0037 & 1.0000 & 0.0003 & 0.0028 & 1.0000 \\
\midrule
\textbf{Combined} & 3.5824 & 3.5757 & 7.8779 & 7.8613 & 10.6605 & 10.5945 & 0.0005 & 0.1935 & 0.9999 & 0.0006 & 0.0766 & 0.9994 \\
\bottomrule
\end{tabular}
}
\end{table*}

\begin{table*}[t]
\centering
\caption{{\it TripletIris} Cross-Implementation Parity and Biometric Performance (Python vs. IREX X C++)}
\label{tab:tripletiris_parity}
\resizebox{\textwidth}{!}{
\begin{tabular}{lcccccccccccc}
\toprule
\multirow{2}{*}{\textbf{Dataset}} & \multicolumn{2}{c}{\textbf{EER (\%)}} & \multicolumn{2}{c}{\textbf{FNMR (\%) @ 0.1\% FMR}} & \multicolumn{2}{c}{\textbf{FNMR (\%) @ 0.01\% FMR}} & \multicolumn{3}{c}{\textbf{Genuine Scores Parity}} & \multicolumn{3}{c}{\textbf{Impostor Scores Parity}} \\
\cmidrule(lr){2-3} \cmidrule(lr){4-5} \cmidrule(lr){6-7} \cmidrule(lr){8-10} \cmidrule(lr){11-13}
 & \textbf{Python} & \textbf{C++} & \textbf{Python} & \textbf{C++} & \textbf{Python} & \textbf{C++} & \textbf{MAD} & \textbf{Max $\Delta$} & \textbf{$R^2$} & \textbf{MAD} & \textbf{Max $\Delta$} & \textbf{$R^2$} \\
\midrule
CASIA-Iris-Lamp & 0.8005 & 0.7999 & 2.3825 & 2.3799 & 5.7169 & 5.7097 & 0.0013 & 0.0150 & 1.0000 & 0.0011 & 0.0136 & 1.0000 \\
CASIA-Iris-Thousand & 2.8166 & 2.8191 & 13.7426 & 13.7046 & 24.8258 & 24.8519 & 0.0139 & 0.1074 & 0.9999 & 0.0124 & 0.1052 & 1.0000 \\
IITD-Iris & 0.1211 & 0.1933 & 0.2706 & 0.2498 & 3.2265 & 3.1224 & 0.1350 & 0.7167 & 0.9853 & 0.1220 & 0.6885 & 0.9968 \\
ND3D & 0.3512 & 0.3498 & 0.8754 & 0.8754 & 2.5809 & 2.5770 & 0.0017 & 0.0164 & 1.0000 & 0.0015 & 0.0196 & 1.0000 \\
WBPMI & 7.8164 & 7.8130 & 45.2731 & 45.2670 & 62.4939 & 62.4761 & 0.0012 & 0.0118 & 1.0000 & 0.0012 & 0.0194 & 1.0000 \\
Q-FIRE & 0.7439 & 0.7467 & 2.4162 & 2.4162 & 7.4809 & 7.4684 & 0.0013 & 0.0236 & 1.0000 & 0.0012 & 0.0219 & 1.0000 \\
\midrule
\textbf{Combined} & 3.3764 & 3.3759 & 14.4518 & 14.4820 & 24.2809 & 24.1648 & 0.0049 & 0.7167 & 1.0000 & 0.0118 & 0.6885 & 0.9999 \\
\bottomrule
\end{tabular}
}
\end{table*}

\begin{table*}[t]
\centering
\caption{{\it CRYPTS} Cross-Implementation Parity and Biometric Performance (MATLAB vs. IREX X C++)}
\label{tab:crypts_parity}
\resizebox{\textwidth}{!}{
\begin{tabular}{lcccccccccccc}
\toprule
\multirow{2}{*}{\textbf{Dataset}} & \multicolumn{2}{c}{\textbf{EER (\%)}} & \multicolumn{2}{c}{\textbf{FNMR (\%) @ 0.1\% FMR}} & \multicolumn{2}{c}{\textbf{FNMR (\%) @ 0.01\% FMR}} & \multicolumn{3}{c}{\textbf{Genuine Scores Parity}} & \multicolumn{3}{c}{\textbf{Impostor Scores Parity}} \\
\cmidrule(lr){2-3} \cmidrule(lr){4-5} \cmidrule(lr){6-7} \cmidrule(lr){8-10} \cmidrule(lr){11-13}
 & \textbf{MATLAB} & \textbf{C++} & \textbf{MATLAB} & \textbf{C++} & \textbf{MATLAB} & \textbf{C++} & \textbf{MAD} & \textbf{Max $\Delta$} & \textbf{$R^2$} & \textbf{MAD} & \textbf{Max $\Delta$} & \textbf{$R^2$} \\
\midrule
CASIA-Iris-Lamp & 1.8507 & 1.3993 & 4.1057 & 2.6016 & 9.3089 & 5.9350 & 0.0470 & 0.3066 & 0.7602 & 0.0263 & 0.1895 & -0.2777 \\
CASIA-Iris-Thousand & 4.3259 & 3.6642 & 12.5417 & 19.5000 & 18.4583 & 89.8750 & 0.0494 & 0.5225 & 0.6982 & 0.0260 & 0.5177 & -0.7323 \\
IITD-Iris & 0.0510 & 0.0000 & 0.6630 & 0.4420 & 0.7735 & 0.4420 & 0.0356 & 0.3706 & 0.7636 & 0.0199 & 0.0892 & 0.0384 \\
ND3D & 1.0496 & 0.5429 & 2.3457 & 1.7901 & 3.3951 & 2.2222 & 0.0417 & 0.1940 & 0.7926 & 0.0212 & 0.1663 & -0.2664 \\
WBPMI & 12.1359 & 12.5500 & 28.8819 & 26.9541 & 37.6095 & 31.3354 & 0.0347 & 0.2731 & 0.9243 & 0.0203 & 0.1981 & -0.2592 \\
Q-FIRE & 3.3769 & 2.6507 & 5.1852 & 4.8148 & 9.4444 & 9.8148 & 0.0475 & 0.2615 & 0.7318 & 0.0281 & 0.2139 & -0.1157 \\
\midrule
\textbf{Combined} & 5.9196 & 5.9971 & 13.1051 & 13.8472 & 17.3132 & 61.2920 & 0.0430 & 0.5225 & 0.8779 & 0.0239 & 0.5177 & -0.4120 \\
\bottomrule
\end{tabular}
}
\end{table*}

\begin{figure*}[!h]
    \centering
    \begin{subfigure}[b]{0.49\textwidth}
        \centering
        \includegraphics[width=\linewidth]{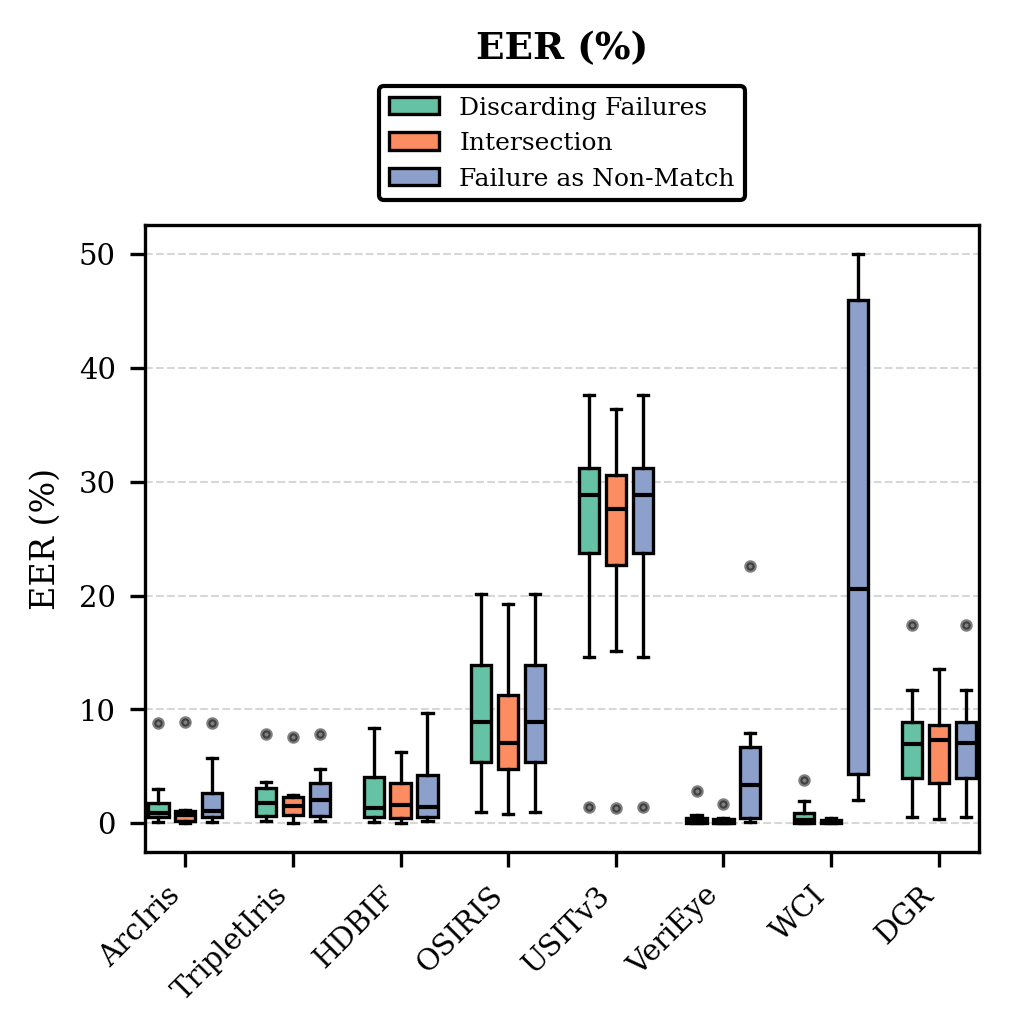}
        \
    \end{subfigure}
    \hfill
    \begin{subfigure}[b]{0.49\textwidth}
        \centering
        \includegraphics[width=\linewidth]{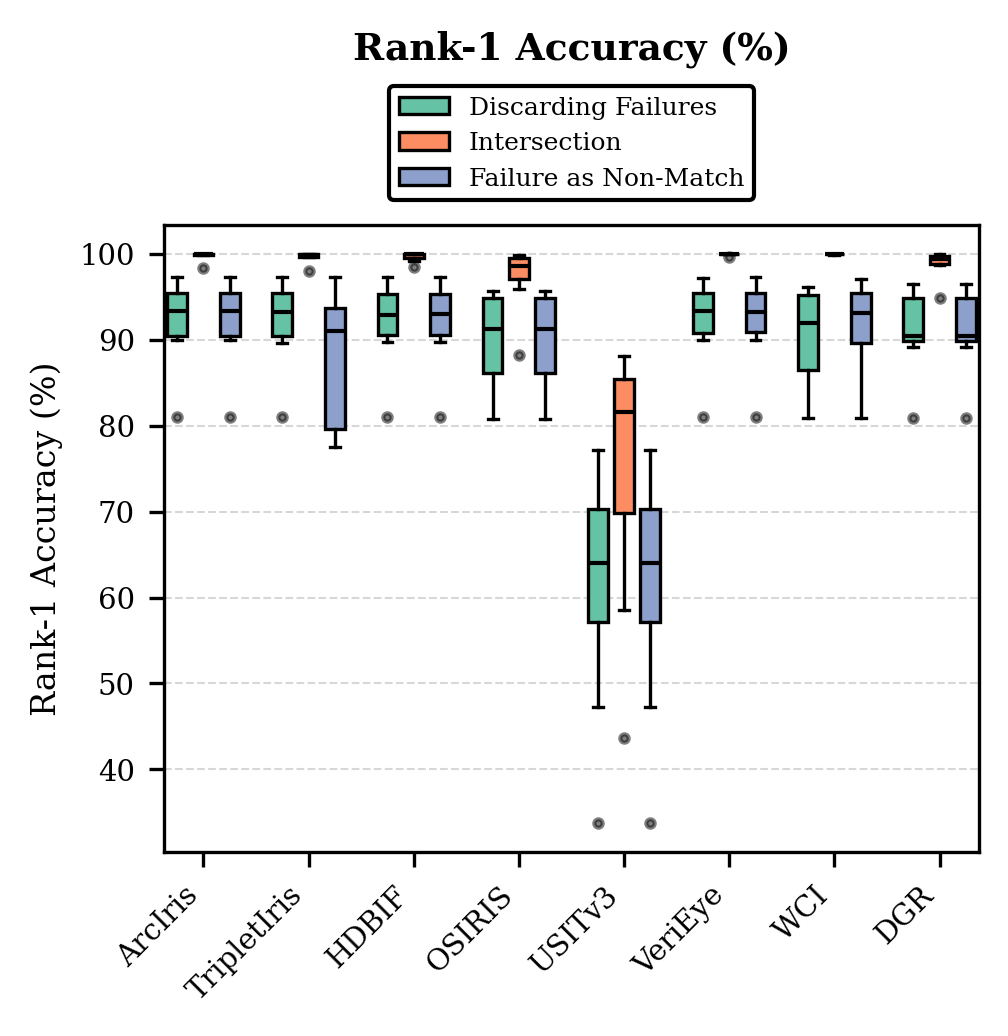}
    \end{subfigure}

    \begin{subfigure}[b]{0.49\textwidth}
        \centering
        \includegraphics[width=\linewidth]{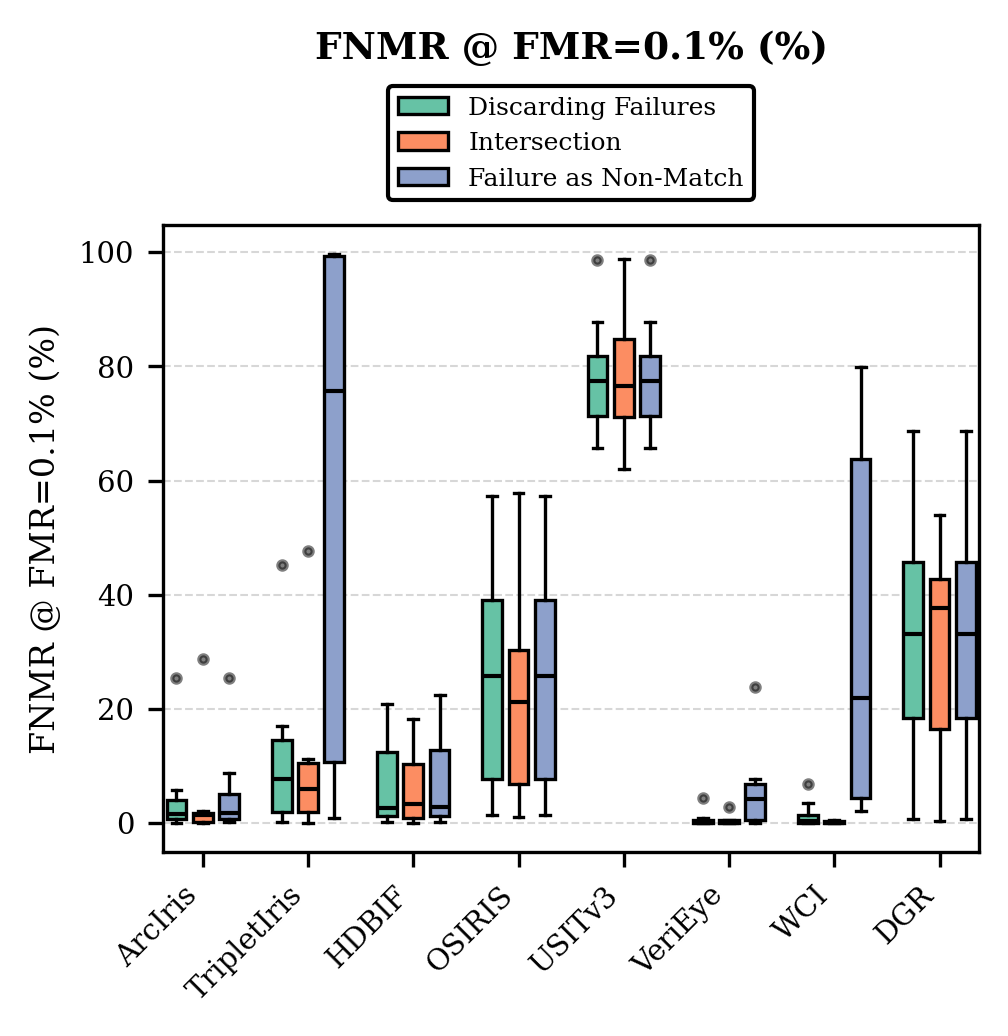}
    \end{subfigure}
    \hfill
    \begin{subfigure}[b]{0.49\textwidth}
        \centering
        \includegraphics[width=\linewidth]{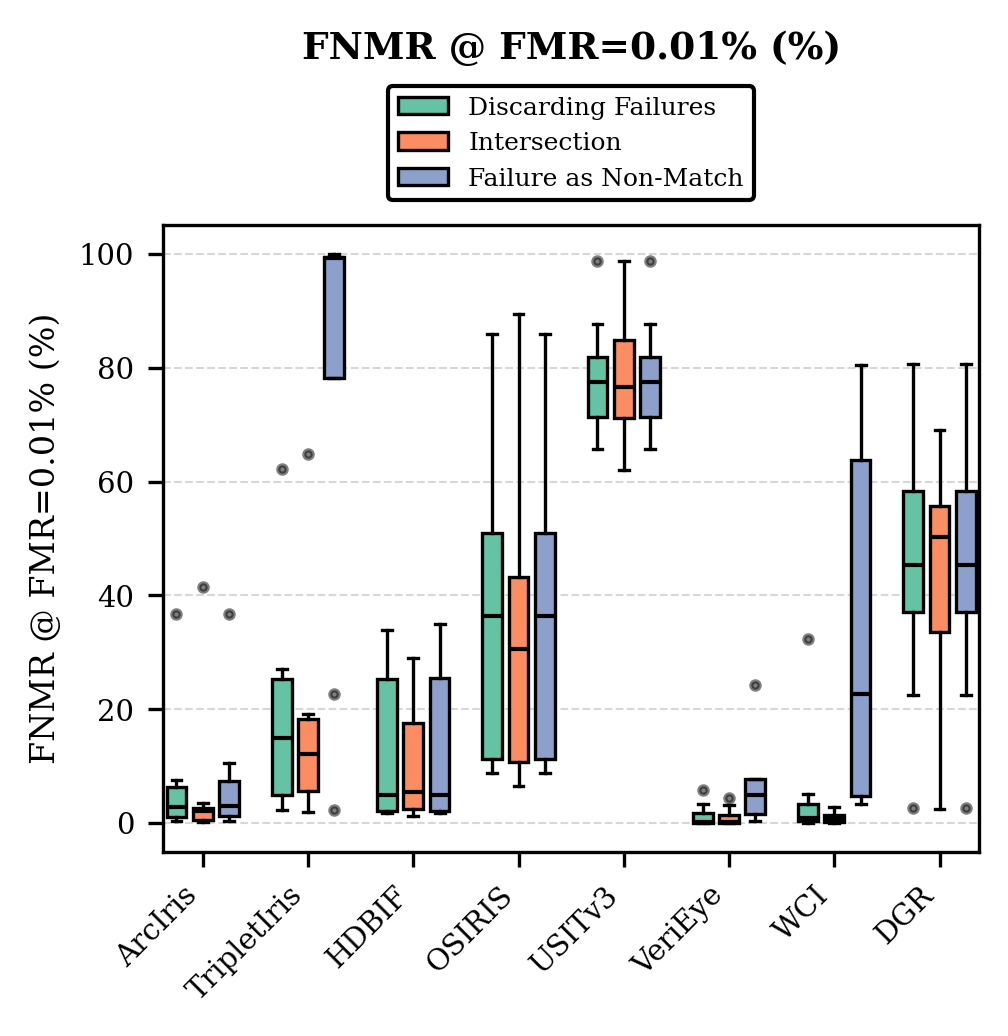}
    \end{subfigure}
    
    \caption{Performance distribution of key metrics (EER, Rank-1 Accuracy, and FNMR at specific operating points) across evaluation conditions: Discarding Failures, Intersection, and treating Failure as Non-Match.}
    \label{fig:comparative_metrics}
\end{figure*}

\section{Performance Evaluation}
\label{sec:performance}

\subsection{IREX X Evaluation and Results}

The IREX X program ranks the algorithms by False Negative Identification Rate (FNIR) at a fixed False Positive Identification Rate (FPIR) equal to 0.01 (1\% false alarm rate). The rank-$k$ hit rate, also reported in IREX, is the fraction of 1:N searches that returned the correct identity within the top $k$ candidates. The uncertainty of FNIR is estimated by providing 90\% confidence intervals. The OPS I-IV sequestered dataset is used in IREX evaluations, which includes 1,000,000 iris images (sourced from 500,000 identities) ``field collected from various locations,'' which ``tend to suffer more from quality-related problems (\eg motion and focus blur) than images collected in more controlled laboratory settings'' \cite{IREXdatasets}. The IREX X official results (leaderboard with dual-eye and single-eye FNIR-based rankings, along with selected demographics-related analyses for each submission) are published and dynamically updated at the NIST IREX X site \cite{IREXrankings} as the algorithms are being submitted. 

At the time of preparing this manuscript, the {\it ArcIris} (named \verb+ ndcvrl_002+ in the IREX X program, and the second open-source algorithm submitted to IREX) has achieved FNIR = $0.048 \pm 0.001$ @ FPIR = $0.01$ for Two-Eye 1:N identification, 97.62\% of rank-1 hit rate, and 98.22\% rank-10 hit rate. This allowed to rank {\it ArcIris} as the 27th most accurate algorithm out of 35 best algorithms included in the IREX X leaderboard. Since IREX, in addition to ranking by the FNIR accuracy, also puts strong limits on processing times and template size, it's noteworthy that {\it ArcIris} is very efficient in both aspects: it is ranked 6th (out of 35 best algorithms) by the iris template size (8kB), and it's in the middle of the leaderboard when the search time is considered ({\it ArcIris} finishes the search among 1 million irises within 2.5 seconds on an Intel(R) Xeon(R) Gold 6140 CPU @ 2.30GHz machine used by NIST to run the evaluations). The first-ever open-source submission to IREX X, {\it TripleIris}, achieved FNIR = $0.326 \pm 0.003$ @ FPIR = $0.01$ for Two-Eye 1:N identification, 89.77\% of rank-1 hit rate, and 93.54\% rank-10 hit rate. Thus, {\it ArcIris} (second) submission is a significant improvement over the first {\it TripleIris} submission.

It is noteworthy that neither {\it ArcIris} nor {\it TripletIris} does any sophisticated iris image quality checks (besides simple checks listed in Sec. \ref{sec:ImageQualityChecks}), which could potentially help in filtering out bad-quality samples and rank higher in IREX X. This is intentional, since the goal of these open-source implementations was to deliver methods that are (a) ``good enough'' and (b) serve as highly-customizable baselines to derive domain-specific iris recognition approaches. For instance, aggressive triaging of iris samples may make these methods unusable in forensic iris applications (such as post-mortem \cite{Trokielewicz_TIFS_2019} or newborn iris recognition \cite{Bhuiyan_2025_WACV}).

\subsection{Selected Research Iris Recognition Benchmarks}

In addition to IREX X evaluations, we evaluated both the proposed and other open-source (and one commercial) iris recognition solutions across eight research datasets to offer insights into how the proposed methods perform for typical research datasets. The commercial systems hide their internal workings, thus comparing them directly against modern open-source options on the same open-source datasets provides an additional (to IREX) understanding of the state of the field. 

\subsubsection{Bridging the Gap to Commercial Systems}
\label{sec:eval_base}

The results show that the proposed methods, {\it ArcIris} and {\it TripletIris}, establish a new standard for open-source iris recognition. Across the evaluations detailed in Table~\ref{tab:performance_metrics}, these algorithms consistently bridge the gap between the open-source approaches and the IREX X top-ranked commercial approach on research datasets. 
{\it ArcIris} consistently achieves an Area Under the Curve (AUC) exceeding 0.99 on almost all datasets, performing very similarly to the commercial {\it VeriEye} system. 

Furthermore, both {\it ArcIris} and {\it TripletIris} significantly outperform legacy open-source models ({\it OSIRIS}, {\it USITv3}, and {\it DGR}), lowering the EER on challenging datasets like CASIA-Iris-Thousand and WBPMI. Between our two proposed models, {\it ArcIris} demonstrates superior robustness. It is especially effective at minimizing the False Non-Match Rate (FNMR) under strict False Match Rate (FMR) conditions. For instance, at FMR=0.1\% on CASIA-Iris-Thousand, {\it ArcIris} maintains an FNMR of just 3.40\%, compared to 13.70\% for {\it TripletIris} and 33.77\% for {\it OSIRIS}. 

In a closed-set identification scenario, {\it ArcIris} performs exceptionally well, frequently matching or even slightly exceeding commercial systems. On ND3D and VII-Q-R2, {\it ArcIris} achieves rank-1 accuracies of 96.30\% and 95.12\%, respectively, narrowly outperforming {\it VeriEye}. The decidability index ($d'$) \cite{Daugman_PAMI_1993}, measuring the separation between genuine and impostor distributions, 
is frequently larger or marginally smaller for {\it ArcIris} compared to {\it VeriEye} in the intersected set of pairs, as shown in Table~\ref{tab:performance_metrics_intersection}. The $d'$ index assumes that comparison scores are normally distributed, hence it's provided here more for completeness (due to the popularity of $d'$ in iris recognition literature) rather than to make formal comparisons among methods.

\begin{table*}[!htbp]
\centering
\caption{Performance metrics across datasets and methods {\bf when failures to extract the template or to compare templates are ignored}. That is, the failure scores (\eg -1.0 or NaN) are simply discarded, and performance metrics are calculated using the remaining scores independently for each method and dataset combination. The best result in each row is {\bf bolded}.}
\label{tab:performance_metrics}
\begin{tabular}{llcccccccc}
\toprule
\textbf{Dataset} & \textbf{Metric} & \textbf{ArcIris} & \textbf{TripletIris} & \textbf{HDBIF} & \textbf{OSIRIS} & \textbf{USITv3} & \textbf{DGR} & \textbf{VeriEye} & \textbf{WCI} \\
\midrule
CASIA-Iris-Lamp & AUC & 0.9990 & 0.9995 & 0.9969 & 0.9799 & 0.6970 & 0.9816 & \textbf{1.0000} & \textbf{1.0000} \\
 & EER (\%) & 0.70 & 0.80 & 1.43 & 6.00 & 27.43 & 6.38 & \textbf{0.00} & 0.22 \\
 & d' & 4.92 & 4.91 & 4.48 & 3.03 & 0.34 & 3.11 & 5.42 & \textbf{5.88} \\
 & Rank-1 (\%) & 95.17 & 95.17 & 95.09 & 94.76 & 60.49 & 94.58 & \textbf{95.20} & 95.12 \\
 & Rank-5 (\%) & 95.18 & 95.18 & 95.16 & 95.10 & 92.19 & 95.09 & \textbf{95.20} & 95.13 \\
 & FNMR @ FMR=0.1\% & 1.19 & 2.38 & 3.11 & 17.89 & 76.26 & 34.90 & \textbf{0.00} & 0.29 \\
 & FNMR @ FMR=0.01\% & 1.82 & 5.64 & 6.29 & 24.77 & 76.26 & 47.87 & \textbf{0.00} & 0.53 \\
 & FTE (\%) & 0.62 & 0.62 & 0.26 & \textbf{0.00} & \textbf{0.00} & 0.01 & 0.51 & 3.34 \\
\midrule
CASIA-Iris-Thousand & AUC & 0.9984 & 0.9967 & 0.9930 & 0.9380 & 0.6985 & 0.9906 & \textbf{1.0000} & 0.9988 \\
 & EER (\%) & 1.32 & 2.82 & 3.26 & 11.75 & 26.83 & 4.19 & \textbf{0.00} & 0.57 \\
 & d' & 3.71 & 3.82 & 3.32 & 2.20 & 0.28 & 3.32 & \textbf{4.61} & 4.46 \\
 & Rank-1 (\%) & 89.96 & 89.69 & 89.77 & 87.07 & 33.73 & 89.21 & \textbf{90.00} & 89.38 \\
 & Rank-5 (\%) & \textbf{90.06} & 90.03 & 90.02 & 88.75 & 52.05 & 89.96 & 90.00 & 89.40 \\
 & FNMR @ FMR=0.1\% & 3.40 & 13.70 & 10.77 & 33.77 & 78.70 & 20.35 & \textbf{0.00} & 0.80 \\
 & FNMR @ FMR=0.01\% & 5.82 & 24.85 & 24.43 & 47.96 & 78.70 & 41.83 & \textbf{0.00} & 1.14 \\
 & FTE (\%) & 0.60 & 0.60 & 0.43 & \textbf{0.00} & \textbf{0.00} & \textbf{0.00} & 2.35 & 12.18 \\
\midrule
IITD-Iris & AUC & \textbf{1.0000} & \textbf{1.0000} & 0.9999 & 0.9986 & 0.9820 & 0.9979 & \textbf{1.0000} & \textbf{1.0000} \\
 & EER (\%) & 0.08 & 0.21 & 0.14 & 1.00 & 1.46 & 0.52 & 0.04 & \textbf{0.02} \\
 & d' & \textbf{12.28} & 6.97 & 10.87 & 6.16 & 4.88 & 9.15 & 8.48 & 12.06 \\
 & Rank-1 (\%) & 81.00 & \textbf{81.04} & 81.00 & 80.78 & 67.18 & 80.92 & \textbf{81.04} & 80.88 \\
 & Rank-5 (\%) & \textbf{81.04} & \textbf{81.04} & \textbf{81.04} & 81.01 & 80.96 & 81.01 & \textbf{81.04} & 80.88 \\
 & FNMR @ FMR=0.1\% & 0.08 & 0.25 & 0.17 & 1.37 & 98.71 & 0.71 & \textbf{0.00} & \textbf{0.00} \\
 & FNMR @ FMR=0.01\% & \textbf{0.79} & 2.83 & 1.87 & 9.75 & 98.71 & 2.64 & 3.35 & 2.80 \\
 & FTE (\%) & 0.12 & 0.12 & 0.12 & \textbf{0.00} & \textbf{0.00} & \textbf{0.00} & 0.12 & 2.11 \\
\midrule
ND3D & AUC & 0.9999 & 0.9999 & 0.9998 & 0.9706 & 0.6856 & 0.9926 & \textbf{1.0000} & \textbf{1.0000} \\
 & EER (\%) & 0.17 & 0.35 & 0.37 & 5.50 & 30.18 & 3.47 & \textbf{0.00} & 0.05 \\
 & d' & 6.52 & 5.33 & 4.96 & 3.56 & 0.37 & 3.61 & 6.17 & \textbf{6.55} \\
 & Rank-1 (\%) & \textbf{96.30} & \textbf{96.30} & \textbf{96.30} & 95.74 & 60.87 & 95.93 & 96.29 & 96.20 \\
 & Rank-5 (\%) & \textbf{96.30} & \textbf{96.30} & \textbf{96.30} & 96.23 & 85.86 & \textbf{96.30} & 96.29 & 96.20 \\
 & FNMR @ FMR=0.1\% & 0.22 & 0.87 & 0.76 & 7.60 & 72.16 & 12.60 & \textbf{0.00} & 0.05 \\
 & FNMR @ FMR=0.01\% & 0.37 & 2.40 & 1.81 & 11.69 & 72.16 & 22.52 & \textbf{0.00} & 0.08 \\
 & FTE (\%) & \textbf{0.00} & \textbf{0.00} & \textbf{0.00} & \textbf{0.00} & \textbf{0.00} & \textbf{0.00} & 0.52 & 5.12 \\
\midrule
WBPMI & AUC & 0.9653 & 0.9762 & 0.9751 & 0.8672 & 0.6768 & 0.9726 & \textbf{0.9926} & 0.9830 \\
 & EER (\%) & 8.83 & 7.81 & 6.58 & 20.16 & 32.94 & 7.62 & \textbf{2.87} & 3.79 \\
 & d' & 2.09 & 2.70 & 2.59 & 1.47 & 0.44 & 2.23 & 1.84 & \textbf{2.75} \\
 & Rank-1 (\%) & 90.61 & 90.68 & 90.88 & 83.41 & 47.29 & 90.72 & \textbf{91.13} & 82.89 \\
 & Rank-5 (\%) & 91.13 & \textbf{91.27} & 91.26 & 86.51 & 66.30 & 91.24 & 91.24 & 83.59 \\
 & FNMR @ FMR=0.1\% & 25.50 & 45.27 & 17.74 & 57.35 & 69.07 & 31.22 & \textbf{4.41} & 6.89 \\
 & FNMR @ FMR=0.01\% & 36.72 & 62.18 & 28.19 & 85.85 & 69.07 & 42.98 & \textbf{5.75} & 32.28 \\
 & FTE (\%) & 0.06 & 0.06 & \textbf{0.00} & \textbf{0.00} & \textbf{0.00} & \textbf{0.00} & 2.54 & 85.91 \\
\midrule
Q-FIRE & AUC & 0.9987 & 0.9994 & 0.9979 & 0.9083 & 0.6216 & 0.9746 & \textbf{0.9999} & 0.9991 \\
 & EER (\%) & 0.68 & 0.74 & 1.16 & 16.45 & 37.65 & 7.96 & \textbf{0.13} & 0.40 \\
 & d' & 5.64 & 4.85 & 4.93 & 1.95 & 0.24 & 2.78 & 5.61 & \textbf{6.57} \\
 & Rank-1 (\%) & 97.26 & \textbf{97.32} & 97.29 & 95.02 & 77.15 & 96.55 & 97.21 & 95.36 \\
 & Rank-5 (\%) & 97.31 & \textbf{97.32} & \textbf{97.32} & 96.81 & 90.53 & 97.24 & 97.21 & 95.39 \\
 & FNMR @ FMR=0.1\% & 0.93 & 2.42 & 2.19 & 39.16 & 65.63 & 43.74 & \textbf{0.13} & 0.52 \\
 & FNMR @ FMR=0.01\% & 1.27 & 7.37 & 3.53 & 49.46 & 65.63 & 62.12 & \textbf{0.15} & 0.69 \\
 & FTE (\%) & 0.04 & 0.04 & 0.04 & \textbf{0.00} & \textbf{0.00} & \textbf{0.00} & 7.74 & 66.53 \\
\midrule
IIITD-CLI & AUC & 0.9972 & 0.9964 & 0.9992 & 0.9760 & 0.8146 & 0.9517 & 0.9999 & \textbf{1.0000} \\
 & EER (\%) & 1.01 & 2.93 & 0.60 & 4.97 & 14.59 & 11.70 & 0.35 & \textbf{0.03} \\
 & d' & 4.09 & 3.68 & 4.60 & 3.49 & 0.86 & 2.25 & 4.68 & \textbf{7.32} \\
 & Rank-1 (\%) & \textbf{91.63} & 91.58 & \textbf{91.63} & 91.13 & 70.13 & 90.23 & 91.57 & 87.70 \\
 & Rank-5 (\%) & \textbf{91.68} & 91.63 & \textbf{91.68} & 91.51 & 91.12 & \textbf{91.68} & 91.57 & 87.70 \\
 & FNMR @ FMR=0.1\% & 2.00 & 13.00 & 1.39 & 7.82 & 87.70 & 51.66 & 0.39 & \textbf{0.00} \\
 & FNMR @ FMR=0.01\% & 3.72 & 22.63 & 2.25 & 8.86 & 87.70 & 57.09 & 0.40 & \textbf{0.00} \\
 & FTE (\%) & \textbf{0.00} & \textbf{0.00} & \textbf{0.00} & \textbf{0.00} & \textbf{0.00} & \textbf{0.00} & 6.77 & 63.08 \\
\midrule
VII-Q-R2 & AUC & 0.9898 & 0.9930 & 0.9606 & 0.9284 & 0.6785 & 0.9000 & \textbf{0.9984} & 0.9955 \\
 & EER (\%) & 3.01 & 3.63 & 8.39 & 13.02 & 30.57 & 17.40 & \textbf{0.70} & 1.91 \\
 & d' & 3.64 & 3.65 & 2.66 & 2.15 & 0.27 & 1.67 & 3.31 & \textbf{3.68} \\
 & Rank-1 (\%) & \textbf{95.12} & 94.91 & 94.22 & 91.43 & 70.73 & 90.14 & 95.08 & 94.58 \\
 & Rank-5 (\%) & \textbf{95.48} & \textbf{95.48} & 95.12 & 94.35 & 88.74 & 94.23 & 95.13 & 94.77 \\
 & FNMR @ FMR=0.1\% & 5.80 & 16.98 & 20.93 & 39.00 & 79.88 & 68.61 & \textbf{0.94} & 3.53 \\
 & FNMR @ FMR=0.01\% & 7.66 & 27.07 & 33.87 & 55.24 & 79.88 & 80.58 & \textbf{1.36} & 5.10 \\
 & FTE (\%) & 2.90 & 2.90 & 1.77 & \textbf{0.00} & \textbf{0.00} & \textbf{0.00} & 22.81 & 30.86 \\
\bottomrule
\end{tabular}
\end{table*}

\subsubsection{Robustness and the Impact of FTE}
\label{sec:eval_FTE_as_errors}

The biggest differences among the methods can be seen in a real-world scenario, in which FTE errors add to system errors, as recommended by ISO/IEC 19794-1. That is, failures to extract or compare templates are treated as non-match outcomes, which are correct when impostor samples are compared, but are incorrect when genuine samples are compared. These results are shown in Table~\ref{tab:performance_metrics_failure_as_nonmatch}. 

Often, methods are accurate only after removing challenging images from processing and comparisons. For example, WCI does well on perfect data but fails to process a large amount of challenging data, resulting in a Failure to Enroll (FTE) rate of 85.91\% on the WBPMI dataset, and FTE=66.53\% on Q-FIRE. When these failures are treated as non-matches, the WCI's EER grows significantly to 45.08\% and 48.74\%, respectively (\ie the classification accuracy close to random chance).

It is interesting to see that {\it VeriEye} also filters out difficult images more aggressively, failing to process 22.81\% of the samples on the VII-Q-R2 dataset and 7.74\% on Q-FIRE. Given that at the time of preparing this manuscript the {\it VeriEye} method is ranked first in the IREX X program (listed as `neurotechnology\_020` in the leaderboard), this may suggest that more aggressive quality control contributes to a good ranking in the NIST program.

In contrast, {\it ArcIris} and {\it TripletIris} maintain near-zero FTE rates (such as 0.06\% on WBPMI and 0.04\% on Q-FIRE), and maintain competitive accuracy when extracting useful identity features from poor-quality, off-angle, or partially covered iris images. Because these newly proposed methods do not artificially boost their accuracy by ignoring hard cases, they may be more suited for real-world deployments where image quality will vary.

\begin{table*}[!htbp]
\centering
\caption{Same as in Table \ref{tab:performance_metrics}, except that {\bf failures to extract the template or to compare templates are treated as non-match errors (real-world scenario)}. To treat failures as non-matches, we replace the failure scores (-1.0 and NaN) with the maximum possible impostor score (for methods using dissimilarity as the comparison score, specifically, namely $\pi$ for {\it ArcIris}, 10,000 for {\it TripletIris}, and 1.0 for Hamming distance-based matchers: {\it HDBIF}, {\it OSIRIS}, {\it USITv3} and {\it WCI}) and with 0.0 (for methods using similarity as the comparison score: {\it DGR} and {\it VeriEye}).}
\label{tab:performance_metrics_failure_as_nonmatch}
\begin{tabular}{llcccccccc}
\toprule
\textbf{Dataset} & \textbf{Metric} & \textbf{ArcIris} & \textbf{TripletIris} & \textbf{HDBIF} & \textbf{OSIRIS} & \textbf{USITv3} & \textbf{DGR} & \textbf{VeriEye} & \textbf{WCI} \\
\midrule
CASIA-Iris-Lamp & EER (\%) & 1.16 & 1.13 & 1.60 & 6.00 & 27.43 & 6.39 & \textbf{0.41} & 2.88 \\
 & Rank-1 (\%) & 95.18 & 92.82 & 95.09 & 94.76 & 60.49 & 94.58 & \textbf{95.20} & 95.14 \\
 & Rank-5 (\%) & 95.19 & 95.20 & 95.16 & 95.10 & 92.19 & 95.10 & \textbf{95.21} & 95.17 \\
 & FNMR @ FMR=0.1\% & 1.74 & 99.44 & 3.31 & 17.89 & 76.26 & 34.91 & \textbf{0.41} & 3.12 \\
 & FNMR @ FMR=0.01\% & 2.37 & 99.44 & 6.49 & 24.77 & 76.26 & 47.87 & \textbf{0.41} & 3.35 \\
 & FTE (\%) & 0.62 & 0.62 & 0.26 & \textbf{0.00} & \textbf{0.00} & 0.01 & 0.51 & 3.34 \\
\midrule
CASIA-Iris-Thousand & EER (\%) & \textbf{1.69} & 3.07 & 3.51 & 11.75 & 26.83 & 4.19 & 1.87 & 11.31 \\
 & Rank-1 (\%) & 89.98 & 77.57 & 89.77 & 87.07 & 33.73 & 89.21 & 90.02 & \textbf{90.04} \\
 & Rank-5 (\%) & \textbf{90.07} & 90.06 & 90.02 & 88.75 & 52.05 & 89.96 & 90.02 & 90.06 \\
 & FNMR @ FMR=0.1\% & 3.90 & 99.43 & 11.09 & 33.77 & 78.70 & 20.35 & \textbf{1.88} & 11.79 \\
 & FNMR @ FMR=0.01\% & 6.29 & 99.43 & 24.67 & 47.96 & 78.70 & 41.83 & \textbf{1.88} & 12.10 \\
 & FTE (\%) & 0.60 & 0.60 & 0.43 & \textbf{0.00} & \textbf{0.00} & \textbf{0.00} & 2.35 & 12.18 \\
\midrule
IITD-Iris & EER (\%) & 0.10 & 0.23 & 0.19 & 1.00 & 1.41 & 0.51 & \textbf{0.08} & 2.05 \\
 & Rank-1 (\%) & 81.01 & 80.11 & 81.01 & 80.78 & 67.18 & 80.92 & \textbf{81.05} & 80.87 \\
 & Rank-5 (\%) & \textbf{81.05} & \textbf{81.05} & \textbf{81.05} & 81.01 & 80.96 & 81.01 & \textbf{81.05} & 80.87 \\
 & FNMR @ FMR=0.1\% & 0.17 & 99.71 & 0.25 & 1.37 & 98.71 & 0.71 & \textbf{0.08} & 2.06 \\
 & FNMR @ FMR=0.01\% & \textbf{0.85} & 99.71 & 1.96 & 9.75 & 98.71 & 2.64 & 3.43 & 4.35 \\
 & FTE (\%) & 0.12 & 0.12 & 0.12 & \textbf{0.00} & \textbf{0.00} & \textbf{0.00} & 0.12 & 2.11 \\
\midrule
ND3D & EER (\%) & \textbf{0.17} & 0.35 & 0.37 & 5.50 & 30.18 & 3.47 & 0.50 & 4.78 \\
 & Rank-1 (\%) & \textbf{96.30} & \textbf{96.30} & \textbf{96.30} & 95.74 & 60.87 & 95.93 & \textbf{96.30} & 96.28 \\
 & Rank-5 (\%) & \textbf{96.30} & \textbf{96.30} & \textbf{96.30} & 96.23 & 85.86 & \textbf{96.30} & \textbf{96.30} & 96.28 \\
 & FNMR @ FMR=0.1\% & \textbf{0.22} & 0.87 & 0.76 & 7.60 & 72.16 & 12.60 & 0.51 & 4.82 \\
 & FNMR @ FMR=0.01\% & \textbf{0.37} & 2.40 & 1.81 & 11.69 & 72.16 & 22.52 & 0.51 & 4.85 \\
 & FTE (\%) & \textbf{0.00} & \textbf{0.00} & \textbf{0.00} & \textbf{0.00} & \textbf{0.00} & \textbf{0.00} & 0.52 & 5.12 \\
\midrule
WBPMI & EER (\%) & 8.84 & 7.84 & 6.58 & 20.16 & 32.94 & 7.62 & \textbf{4.82} & 45.08 \\
 & Rank-1 (\%) & 90.62 & 90.46 & 90.88 & 83.41 & 47.29 & 90.72 & \textbf{91.16} & 88.22 \\
 & Rank-5 (\%) & 91.14 & \textbf{91.28} & 91.26 & 86.51 & 66.30 & 91.24 & 91.26 & 88.78 \\
 & FNMR @ FMR=0.1\% & 25.52 & 54.51 & 17.74 & 57.35 & 69.07 & 31.22 & \textbf{6.60} & 79.96 \\
 & FNMR @ FMR=0.01\% & 36.73 & 99.01 & 28.19 & 85.85 & 69.07 & 42.98 & \textbf{7.84} & 80.44 \\
 & FTE (\%) & 0.06 & 0.06 & \textbf{0.00} & \textbf{0.00} & \textbf{0.00} & \textbf{0.00} & 2.54 & 85.91 \\
\midrule
Q-FIRE & EER (\%) & \textbf{0.71} & 0.76 & 1.19 & 16.45 & 37.65 & 7.96 & 7.93 & 48.74 \\
 & Rank-1 (\%) & 97.26 & \textbf{97.33} & 97.29 & 95.02 & 77.15 & 96.55 & \textbf{97.33} & 97.12 \\
 & Rank-5 (\%) & 97.31 & \textbf{97.33} & \textbf{97.33} & 96.81 & 90.53 & 97.24 & \textbf{97.33} & 97.19 \\
 & FNMR @ FMR=0.1\% & \textbf{0.96} & 3.65 & 2.23 & 39.16 & 65.63 & 43.74 & 7.75 & 65.60 \\
 & FNMR @ FMR=0.01\% & \textbf{1.30} & 99.97 & 3.56 & 49.46 & 65.63 & 62.12 & 7.77 & 65.67 \\
 & FTE (\%) & 0.04 & 0.04 & 0.04 & \textbf{0.00} & \textbf{0.00} & \textbf{0.00} & 7.74 & 66.53 \\
\midrule
IIITD-CLI & EER (\%) & 1.01 & 2.93 & \textbf{0.60} & 4.97 & 14.59 & 11.70 & 6.34 & 50.00 \\
 & Rank-1 (\%) & \textbf{91.63} & 91.58 & \textbf{91.63} & 91.13 & 70.13 & 90.23 & 91.40 & 91.16 \\
 & Rank-5 (\%) & \textbf{91.68} & 91.63 & \textbf{91.68} & 91.51 & 91.12 & \textbf{91.68} & 91.49 & 91.40 \\
 & FNMR @ FMR=0.1\% & 2.00 & 13.00 & \textbf{1.39} & 7.82 & 87.70 & 51.66 & 6.54 & 63.09 \\
 & FNMR @ FMR=0.01\% & 3.72 & 22.63 & \textbf{2.25} & 8.86 & 87.70 & 57.09 & 6.56 & 63.09 \\
 & FTE (\%) & \textbf{0.00} & \textbf{0.00} & \textbf{0.00} & \textbf{0.00} & \textbf{0.00} & \textbf{0.00} & 6.77 & 63.08 \\
\midrule
VII-Q-R2 & EER (\%) & 5.70 & \textbf{4.75} & 9.71 & 13.02 & 30.57 & 17.40 & 22.62 & 29.92 \\
 & Rank-1 (\%) & \textbf{95.15} & 78.07 & 94.27 & 91.43 & 70.73 & 90.14 & \textbf{95.15} & 95.01 \\
 & Rank-5 (\%) & \textbf{95.51} & 95.37 & 95.17 & 94.35 & 88.74 & 94.23 & 95.21 & 95.21 \\
 & FNMR @ FMR=0.1\% & \textbf{8.78} & 96.76 & 22.38 & 39.00 & 79.88 & 68.61 & 23.93 & 32.01 \\
 & FNMR @ FMR=0.01\% & \textbf{10.64} & 96.76 & 35.05 & 55.24 & 79.88 & 80.58 & 24.24 & 33.11 \\
 & FTE (\%) & 2.90 & 2.90 & 1.77 & \textbf{0.00} & \textbf{0.00} & \textbf{0.00} & 22.81 & 30.86 \\
\bottomrule
\end{tabular}
\end{table*}

\subsubsection{Performance on a Level Playing Field}
\label{sec:eval_common}

The goal of the third evaluation, with the results shown in Table~\ref{tab:performance_metrics_intersection}, was to isolate a subset of ``easy'' images that every method was able to process successfully to compare matching accuracy for samples judged by all algorithms as of sufficient quality. However, due to high FTE observed for some combinations of methods and datasets, we would end up with a tiny and trivial intersected set for that particular benchmark. Thus, we utilize an FTE threshold of 30\% to filter out a method-dataset combination to avoid evaluations on tiny sets. This only removed the WCI method and only for WBPMI, Q-FIRE, IIITD-CLI, and VII-Q-R2 datasets. Under these perfect conditions, the performance gap between the methods becomes negligible. For instance, {\it ArcIris} still stands out, achieving a 100\% rank-1 accuracy on the CASIA-Iris-Lamp, IITD-Iris, ND3D, and Q-FIRE datasets. This suggests that open-source iris recognition is already very strong for good-quality samples, and when the differences among the methods start to be seen is the way how these methods process more challenging, potentially non-ISO/IEC-19794-6-compliant images. 

\begin{table*}[!htbp]
\centering
\caption{Same as in Table \ref{tab:performance_metrics}, except that {\bf only iris images that passed the quality check in all methods are considered}. That is, use only the intersection of the accepted pairs across the different methods. To secure statistically-significant results, if a method has an FTE of greater than 30\% for a particular dataset, we do not evaluate it on that dataset and take the intersection of the accepted pairs for the rest of the methods.}
\label{tab:performance_metrics_intersection}
\begin{tabular}{llcccccccc}
\toprule
\textbf{Dataset} & \textbf{Metric} & \textbf{ArcIris} & \textbf{TripletIris} & \textbf{HDBIF} & \textbf{OSIRIS} & \textbf{USITv3} & \textbf{DGR} & \textbf{VeriEye} & \textbf{WCI} \\
\midrule
CASIA-Iris-Lamp & AUC & 0.9998 & 0.9998 & 0.9981 & 0.9787 & 0.7105 & 0.9814 & \textbf{1.0000} & \textbf{1.0000} \\
 & EER (\%) & 0.28 & 0.84 & 1.35 & 6.21 & 26.44 & 6.97 & \textbf{0.00} & 0.16 \\
 & d' & 5.19 & 4.83 & 4.46 & 3.05 & 0.42 & 2.95 & 5.32 & \textbf{5.59} \\
 & Rank-1 (\%) & \textbf{100.00} & 99.88 & 99.88 & 99.59 & 84.74 & 99.65 & \textbf{100.00} & 99.94 \\
 & Rank-5 (\%) & \textbf{100.00} & \textbf{100.00} & \textbf{100.00} & 99.76 & 98.88 & \textbf{100.00} & \textbf{100.00} & \textbf{100.00} \\
 & FNMR @ FMR=0.1\% & 0.37 & 2.96 & 3.31 & 16.91 & 76.59 & 37.83 & \textbf{0.00} & 0.21 \\
 & FNMR @ FMR=0.01\% & 0.59 & 6.58 & 5.39 & 22.39 & 76.59 & 50.55 & \textbf{0.00} & 0.39 \\
\midrule
CASIA-Iris-Thousand & AUC & 0.9994 & 0.9980 & 0.9945 & 0.9686 & 0.7170 & 0.9942 & \textbf{1.0000} & 0.9991 \\
 & EER (\%) & 0.83 & 2.23 & 2.84 & 7.86 & 25.17 & 3.67 & \textbf{0.00} & 0.47 \\
 & d' & 4.10 & 4.05 & 3.42 & 2.60 & 0.31 & 3.51 & \textbf{4.70} & 4.66 \\
 & Rank-1 (\%) & 99.86 & 99.66 & 99.59 & 98.07 & 43.63 & 99.17 & \textbf{100.00} & 99.93 \\
 & Rank-5 (\%) & \textbf{100.00} & 99.93 & 99.93 & 99.31 & 59.55 & 99.93 & \textbf{100.00} & \textbf{100.00} \\
 & FNMR @ FMR=0.1\% & 2.08 & 10.27 & 9.22 & 25.72 & 82.47 & 18.04 & \textbf{0.00} & 0.65 \\
 & FNMR @ FMR=0.01\% & 3.47 & 19.17 & 15.38 & 38.93 & 82.47 & 37.56 & \textbf{0.00} & 0.92 \\
\midrule
IITD-Iris & AUC & \textbf{1.0000} & \textbf{1.0000} & \textbf{1.0000} & 0.9992 & 0.9839 & 0.9999 & \textbf{1.0000} & \textbf{1.0000} \\
 & EER (\%) & \textbf{0.02} & 0.06 & \textbf{0.02} & 0.81 & 1.36 & 0.36 & \textbf{0.02} & \textbf{0.02} \\
 & d' & \textbf{12.84} & 7.07 & 11.30 & 6.29 & 5.04 & 9.77 & 8.72 & 12.06 \\
 & Rank-1 (\%) & \textbf{100.00} & \textbf{100.00} & \textbf{100.00} & 99.83 & 83.49 & \textbf{100.00} & \textbf{100.00} & \textbf{100.00} \\
 & Rank-5 (\%) & \textbf{100.00} & \textbf{100.00} & \textbf{100.00} & \textbf{100.00} & 99.94 & \textbf{100.00} & \textbf{100.00} & \textbf{100.00} \\
 & FNMR @ FMR=0.1\% & \textbf{0.00} & 0.04 & \textbf{0.00} & 1.15 & 98.77 & 0.47 & \textbf{0.00} & \textbf{0.00} \\
 & FNMR @ FMR=0.01\% & \textbf{0.66} & 2.59 & 1.81 & 10.04 & 98.77 & 2.42 & 3.12 & 2.80 \\
\midrule
ND3D & AUC & \textbf{1.0000} & \textbf{1.0000} & 0.9999 & 0.9718 & 0.6833 & 0.9936 & \textbf{1.0000} & \textbf{1.0000} \\
 & EER (\%) & 0.08 & 0.30 & 0.30 & 5.30 & 30.46 & 3.29 & \textbf{0.00} & 0.02 \\
 & d' & \textbf{6.73} & 5.38 & 5.04 & 3.63 & 0.35 & 3.64 & 6.23 & 6.64 \\
 & Rank-1 (\%) & \textbf{100.00} & \textbf{100.00} & \textbf{100.00} & 99.55 & 73.58 & 99.75 & \textbf{100.00} & \textbf{100.00} \\
 & Rank-5 (\%) & \textbf{100.00} & \textbf{100.00} & \textbf{100.00} & 99.95 & 93.41 & \textbf{100.00} & \textbf{100.00} & \textbf{100.00} \\
 & FNMR @ FMR=0.1\% & 0.07 & 0.69 & 0.56 & 7.18 & 73.31 & 12.10 & \textbf{0.00} & 0.01 \\
 & FNMR @ FMR=0.01\% & 0.15 & 1.90 & 1.28 & 10.92 & 73.31 & 21.84 & \textbf{0.00} & 0.02 \\
\midrule
WBPMI & AUC & 0.9665 & 0.9785 & 0.9782 & 0.8772 & 0.7049 & 0.9740 & \textbf{0.9975} & - \\
 & EER (\%) & 8.86 & 7.61 & 6.21 & 19.23 & 30.79 & 7.66 & \textbf{1.71} & - \\
 & d' & 2.07 & \textbf{2.75} & 2.62 & 1.48 & 0.59 & 2.15 & 1.86 & - \\
 & Rank-1 (\%) & 98.35 & 98.03 & 98.47 & 88.27 & 58.55 & 98.80 & \textbf{99.64} & - \\
 & Rank-5 (\%) & 99.52 & 99.56 & 99.44 & 93.57 & 81.85 & 99.72 & \textbf{99.92} & - \\
 & FNMR @ FMR=0.1\% & 28.74 & 47.63 & 18.26 & 57.75 & 62.04 & 37.45 & \textbf{2.85} & - \\
 & FNMR @ FMR=0.01\% & 41.51 & 64.81 & 28.95 & 89.45 & 62.04 & 50.04 & \textbf{4.44} & - \\
\midrule
Q-FIRE & AUC & 0.9982 & 0.9983 & 0.9963 & 0.8970 & 0.6328 & 0.9749 & \textbf{0.9996} & - \\
 & EER (\%) & 1.14 & 0.81 & 1.89 & 17.85 & 36.38 & 7.61 & \textbf{0.47} & - \\
 & d' & \textbf{5.24} & 4.94 & 4.64 & 1.80 & 0.28 & 2.82 & 5.02 & - \\
 & Rank-1 (\%) & \textbf{100.00} & \textbf{100.00} & \textbf{100.00} & 95.89 & 87.45 & 99.69 & \textbf{100.00} & - \\
 & Rank-5 (\%) & \textbf{100.00} & \textbf{100.00} & \textbf{100.00} & 98.84 & 97.99 & \textbf{100.00} & \textbf{100.00} & - \\
 & FNMR @ FMR=0.1\% & 1.64 & 2.40 & 3.54 & 44.33 & 64.43 & 40.29 & \textbf{0.50} & - \\
 & FNMR @ FMR=0.01\% & 2.36 & 6.91 & 5.43 & 54.61 & 64.43 & 57.12 & \textbf{0.52} & - \\
\midrule
IIITD-CLI & AUC & 0.9996 & 0.9976 & 0.9996 & 0.9830 & 0.8231 & 0.9559 & \textbf{1.0000} & - \\
 & EER (\%) & 0.68 & 2.50 & 0.46 & 3.29 & 15.14 & 11.44 & \textbf{0.00} & - \\
 & d' & 4.30 & 3.79 & 4.65 & 3.87 & 0.90 & 2.28 & \textbf{4.85} & - \\
 & Rank-1 (\%) & \textbf{100.00} & 99.86 & \textbf{100.00} & 99.17 & 88.09 & 98.75 & \textbf{100.00} & - \\
 & Rank-5 (\%) & \textbf{100.00} & \textbf{100.00} & \textbf{100.00} & 99.86 & 99.86 & \textbf{100.00} & \textbf{100.00} & - \\
 & FNMR @ FMR=0.1\% & 1.47 & 11.27 & 1.14 & 5.63 & 91.76 & 50.12 & \textbf{0.00} & - \\
 & FNMR @ FMR=0.01\% & 2.37 & 18.07 & 2.80 & 6.50 & 91.76 & 55.27 & \textbf{0.00} & - \\
\midrule
VII-Q-R2 & AUC & 0.9969 & 0.9967 & 0.9805 & 0.9543 & 0.7146 & 0.9339 & \textbf{0.9992} & - \\
 & EER (\%) & 1.00 & 2.27 & 5.69 & 9.09 & 28.79 & 13.52 & \textbf{0.37} & - \\
 & d' & \textbf{5.28} & 4.06 & 3.15 & 2.52 & 0.44 & 1.95 & 3.32 & - \\
 & Rank-1 (\%) & 99.87 & 99.61 & 99.21 & 97.44 & 79.59 & 94.82 & \textbf{100.00} & - \\
 & Rank-5 (\%) & \textbf{100.00} & \textbf{100.00} & 99.87 & 99.15 & 94.82 & 98.36 & \textbf{100.00} & - \\
 & FNMR @ FMR=0.1\% & 1.49 & 9.18 & 13.65 & 25.40 & 76.60 & 53.95 & \textbf{0.52} & - \\
 & FNMR @ FMR=0.01\% & 2.04 & 17.41 & 24.50 & 39.44 & 76.60 & 69.12 & \textbf{0.91} & - \\
\bottomrule
\end{tabular}
\end{table*}

\subsubsection{Commercial vs. Open-Source Paradigm} 

It is important to acknowledge that {\it VeriEye}, a mature commercial product, yields the lowest baseline EERs and highest $d'$ values across most research datasets. However, 
as a proprietary ``black box'' system, there is no 
publicly available information about what data was used to design this algorithm. For instance, it is possible that there is a significant overlap between the data used to train the {\it VeriEye} models and the test datasets used in evaluations provided in this paper. If this ``data leakage'' exists, the performance metrics associated with the commercial method would be skewed towards more favorable numbers. 

\subsubsection{Summary} 

Fig.~\ref{fig:comparative_metrics} summarizes three key metrics: EER, rank-1 accuracy, and FNMR across our three evaluation conditions: when the low-quality images are ignored (Sec. \ref{sec:eval_base}), when failures add to the system reject/accept metrics (Sec.
\ref{sec:eval_FTE_as_errors}), and when only the same ``easy'' samples for all methods are used (Sec. \ref{sec:eval_common}). It is clear from these plots that the proposed open-source methods provide comparable performance to the commercial method {\it VeriEye}, especially when image processing failures are regarded as non-matches, while outperforming the other existing open-source methods.

\section{Conclusions}

Iris recognition, owing to its maturity, has become the third (along with face and fingerprint recognition) principal biometric technique being integrated into systems influencing national safety. Part of this integration is also an ongoing discussion on creating procedures for human examiners of iris images, who might be called upon to give testimony in court, and may require using open-source iris recognition algorithms that are transparent, interpretable, and vetted in a trusted program with large-scale, sequestered and variable-quality data. NIST IREX is a well-established platform for independent evaluation of iris recognition algorithms and publishes the performance analyses for top iris recognition solutions. However, there were no submissions of open-source algorithms to IREX programs to date, and thus, a comprehensive assessment of where the open-source iris recognition community stands in terms of performance, compared to the leading commercial solutions, was not yet possible on a large scale. 

This paper delivers the first IREX-compliant (written in C++ and wrapped with the newest IREX X API v3.0) open-source versions of one iris segmentation and four iris recognition methods, including two new and very competitive iris image encoding and matching algorithms: {\it ArcIris} and {\it TripletIris}. By making these new methods open-source, we are providing the research community with powerful and independently tested in the NIST IREX program tools that can be trusted in real-world situations. Moreover, as we provide both Python and IREX-compliant C++ implementations, our code and this paper can act as a guideline for researchers interested in submitting their own iris recognition algorithms for evaluation in the IREX X program, lowering the barrier, especially for academic participants. Since the implemented methods can be incorporated into any other iris recognition system, the value of the designed modules potentially extends beyond the IREX X evaluation.

\appendices

\section*{Acknowledgments}
The development of the C++ (IREX X-compliant) versions of the HDBIF and CRYPTS methods was supported by the U.S. Department of Commerce (grant No. 60NANB22D153). The views and conclusions contained in this document are those of the authors and should not be interpreted as representing the official policies, either expressed or implied, of the U.S. Department of Commerce or the U.S. Government.

The authors would like to thank Dr. James Matey at NIST for his encouragement to start and pursue this project, and for his feedback on the initial draft of this manuscript. 

Finally, the authors also extend their gratitude to the NIST IREX team for processing the early submissions and providing code recommendations that ensured the final submissions met the IREX processing time standards.

\ifCLASSOPTIONcaptionsoff
  \newpage
\fi

\FloatBarrier
\bibliographystyle{IEEEtran}
\bibliography{aczajka-refs}

\end{document}